\documentclass[preprint,12pt]{elsarticle}

\usepackage{listings}
\usepackage{amssymb,amsmath}
\usepackage{booktabs}
\usepackage{enumitem}
\usepackage{multicol}
\usepackage[hyphens]{url}
\usepackage{makecell}
\usepackage{soul}
\usepackage{multirow}
\usepackage{graphicx}
\usepackage[table,xcdraw]{xcolor}
\usepackage[section]{placeins}
\usepackage{spverbatim}
\usepackage[margin=1in]{geometry}

\usepackage{fancyvrb}
\usepackage{fvextra}
\usepackage{framed}
\usepackage{caption}
\usepackage{float}
\usepackage{anyfontsize}

\usepackage[normalem]{ulem}
\usepackage{lscape}

\makeatletter
\fvset{listparameters={\setlength{\topsep}{0pt}\setlength{\partopsep}{0pt}\setlength{\parskip}{0pt}\setlength{\parsep}{0pt}}}
\makeatother

\usepackage[commandnameprefix=always,final]{changes}
\definechangesauthor[color=olive]{del}
\definechangesauthor[color=olive]{add}
\definechangesauthor[color=olive]{delr2}
\definechangesauthor[color=olive]{addr2}
\definechangesauthor[color=red]{delr3}
\definechangesauthor[color=red]{addr3}

\makeatletter
\newcommand\medlarge{\@setfontsize\medlarge{13.5}{15}}
\makeatother

\usepackage{listings}
\usepackage{svg}

\usepackage{amsthm}

\journal{Data and Knowledge Engineering}

\begin{document}

\begin{frontmatter}



\title{How well can \chadded[id=addr3]{a} large language model\chdeleted[id=addr3]{s} explain business processes as perceived by users?\tnoteref{t1}} \tnotetext[t1]{This project has received funding from the European Union’s Horizon research and innovation programme under grant agreements no 101092639 (FAME), 101094905 (AI4GOV), and 101092021 (AutoTwin).}



\author[o1]{Dirk Fahland}
\ead{d.fahland@tue.nl}

\author[o2]{Fabiana Fournier}
\ead{fabiana@il.ibm.com}

\author[o2]{Lior Limonad}
\ead{liorli@il.ibm.com}

\author[o2]{Inna Skarbovsky}
\ead{inna@il.ibm.com}

\author[o1]{Ava J.E. Swevels}
\ead{a.j.e.swevels@tue.nl}

\affiliation[o1]{organization={Eindhoven University of Technology},
            city={Eindhoven},
            country={Netherlands}}
\affiliation[o2]{organization={IBM Research},
            city={Haifa},
            country={Israel}}

\begin{abstract}
Large Language Models (LLMs) are trained on a vast amount of text to interpret and generate human-like textual content. They are becoming a vital vehicle in realizing the vision of the autonomous enterprise, with organizations today actively adopting LLMs to automate many aspects of their operations.
LLMs are likely to play a prominent role in future AI-augmented business process management systems (ABPMSs) catering functionalities across all system lifecycle stages. 
One such system's functionality is Situation-Aware eXplainability (SAX), which relates to generating causally sound and yet human-interpretable explanations that take into account the process context in which the explained condition occurred.



In this paper, we present the SAX4BPM framework developed to generate SAX explanations. The SAX4BPM suite consists of a set of services and a central knowledge repository.  
The functionality of these services is to elicit the various knowledge ingredients that underlie SAX explanations.  A key innovative component among these ingredients is the causal process execution view. In this work, we integrate the framework with an LLM to leverage its power to synthesize the various input ingredients for the sake of improved SAX explanations.

Since the use of LLMs for SAX is also accompanied by a certain degree of doubt related to its capacity to adequately fulfill SAX along with its tendency for hallucination and lack of inherent capacity to reason, 
we pursued a methodological evaluation of the \chadded[id=addr2]{perceived} quality of the generated explanations.
To this aim, we developed a designated scale and conducted a rigorous user study. Our findings show that the input presented to the LLMs aided with the guard-railing of its performance, yielding SAX explanations having better-perceived fidelity. This improvement is moderated by the perception of trust and curiosity. More so, this improvement comes at the cost of the perceived interpretability of the explanation.




\end{abstract}



\begin{keyword}



  Business Process \sep
  Methodologies and Tools \sep
  AI \sep
  Explainability \sep
  Large Language Models

\end{keyword}

\end{frontmatter}



\newif\ifshowcomments
\showcommentstrue
\ifshowcomments
\newcommand{\mynote}[2]{\fbox{\bfseries\sffamily\scriptsize{#1}}
{\small$\blacktriangleright$\textsf{#2}$\blacktriangleleft$}}
\else
\newcommand{\mynote}[2]{}
\fi
\newcommand{\inna}[1]{\textcolor{red}{\mynote{Inna}{#1}}}
\newcommand{\lior}[1]{\textcolor{blue}{\mynote{LIOR}{#1}}}
\newcommand{\fabiana}[1]{\textcolor{red}{\mynote{Fabiana}{#1}}}
\newcommand{\dirk}[1]{\textcolor{red}{\mynote{Dirk}{#1}}}
\newcommand{\ava}[1]{\textcolor{red}{\mynote{Ava}{#1}}}
\setlist[itemize]{noitemsep}
\section{Introduction and motivation}
Explanations are the foundation for the adoption and trust of humans in business processes augmented by AI (ABPMSs)~\cite{Dumas2023}. Via explanations, process users can understand and act upon the various situations that evolve during process executions. Thus, explainability is deemed as an inherent functionality of the ABPMS.

Explanations in the context of business processes (BPs) are expected to take advantage of the knowledge of the BP definitions and full runtime process traces. Furthermore, they are expected to embed the ability to go beyond a local reasoning context, handle a large variety of situations, and facilitate the (automatic or by humans) tracking of execution consistency for a better understanding of process flows and process outcomes. Adequate explanations of BPs are key to driving ongoing process improvements.

According to Forrester’s Q3 2022 survey results, process optimization and automation are now mainstream skills that directly align with strategic business objectives\footnote{\url{https://www.forrester.com/report/forresters-q3-2022-digital-process-automation-survey-results-organizations-respond-to-the-automation-imperative/RES178220}}. This progress of automation and AI also presents an opportunity for the automation of explainability in the area of BPM. The global business process management (BPM) industry generated \$15.4 billion in 2022 and is anticipated to generate \$65.8 billion by 2032\footnote{\url{https://www.marketwatch.com/press-release/ business-process-management-bpm-market-to-reach-65-8-billion-globally-by-2032-at-15-8-cagr-allied-market-research-25380466}}.
As with the evolution of technologies for explainability that accompany recent advances in ML, explainability in BP has a great potential to untangle the evolving complexity of organizational processes and promote trust in and adoption of the automation technology (e.g.,~\cite{Zhou2021}).

One way to automate explainability is to exploit Large Language Models (LLMs) capabilities. Concretely, as supported by recent Gartner's report~\cite{Gartner2023}, LLMs are being adopted across all industries and business functions, driving a variety of use cases such as text summarization, question-answering, document translation, and alike. LLMs can also be augmented by additional capabilities to create more powerful systems, and feature a growing ecosystem of tools. Among these, we foresee the benefit of leveraging LLMs as a means for the automation of explanations in BPs. 

\chadded[id=add]{Our work presented in this paper addresses the growing interest in using LLMs as a tool for various data and knowledge-processing tasks. For example, the recent study~\cite{Remadi2024ToData} explored LLMs for data extraction from unstructured sources and entity resolution. Trained on large corpora across different domains, LLMs offer the potential to replace traditional data-driven tools, such as rule conformance checking tools for BPM, as discussed in~\cite{Corradini2018AModels}. We also follow a data-driven approach, using LLMs to synthesize multiple knowledge ingredients to create sound and interpretable explanations about business processes.}

However, the use of LLMs as a vehicle for explanation generation comes with a certain caveat. While LLMs present a remarkable performance in various NLP tasks without the need for supervised training~\cite{Gao2023}, they are not good causal reasoners~\cite{Subbarao2024,Moritz2022} and exhibit a critical tendency to produce hallucinations~\cite{Huang2023}, resulting in content that is inconsistent with real-world facts or user inputs. This phenomenon poses substantial challenges to their practical deployment and raises concerns over the reliability of LLMs in real-world scenarios, which attracts increasing attention to detect and mitigate these hallucinations.

Our main objective is to guardrail the functionality of LLMs such that they can be leveraged to give explanations that \chadded[id=add]{the users grasp as} adequately point\chadded[id=add]{ing} to the circumstances that bring about the unfolding of BPs towards certain conditions. \chadded[id=add]{While we acknowledge that the objective quality of explanations is important, in this work we attend to the use of LLMs as a means to synthesize the explanations by articulating the various system's output contents in a textual form, where the eventual explanations presented alter users' perceptions.}
We hypothesize in this work that the perceived quality of LLM-generated explanations of this sort could be altered and even improved via the methodological injection of different knowledge articulations \chdeleted[id=del]{that are} given to it as an input leveraging LLM prompt engineering. 

\begin{figure}[ht]
    \centering
    \includegraphics[width=\textwidth]{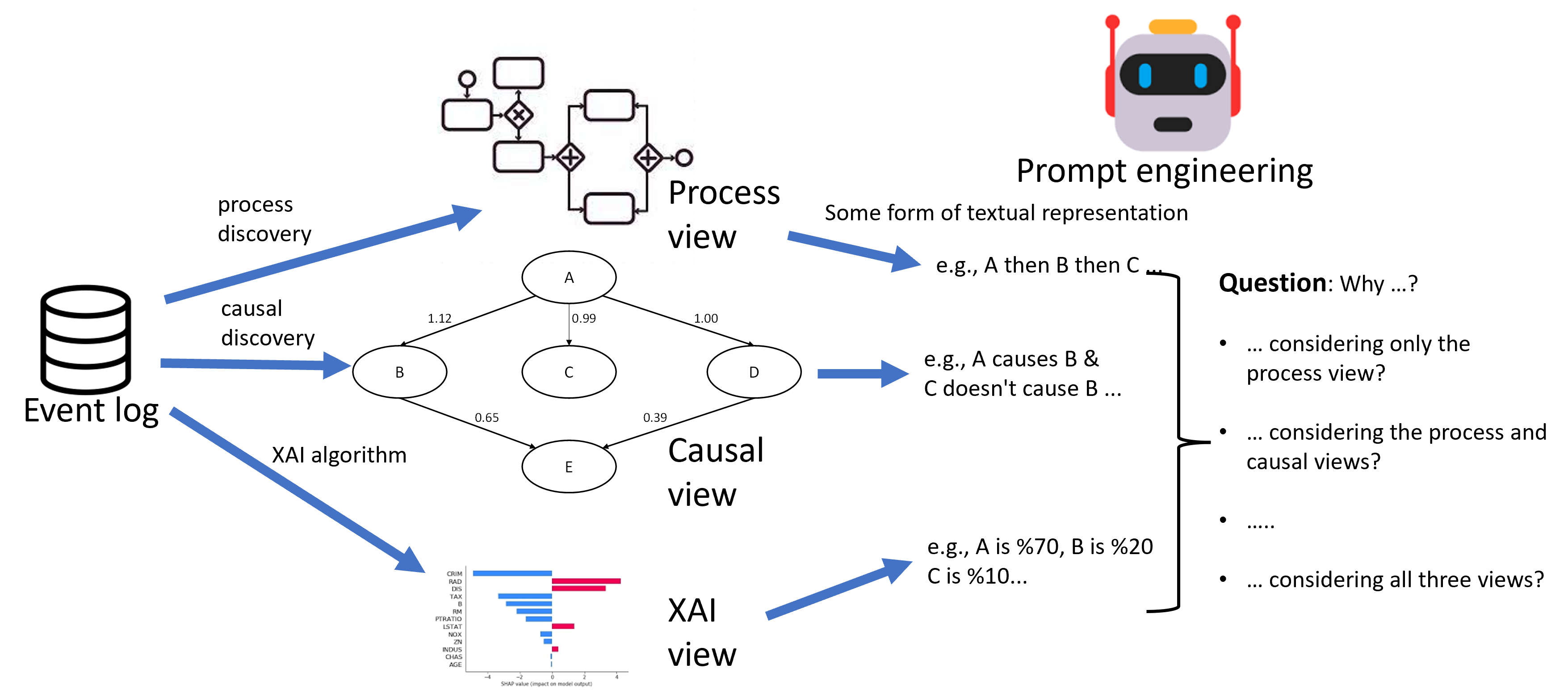}
    \caption{General approach}
    \label{fig:general-approach}
\end{figure}

Furthermore, we focus on the incorporation of a recently developed view that highlights the causal execution dependencies among the activities in the BP~\cite{Fournier2023TheDependencies}. We add it as an additional input to the LLM. To our knowledge, this is the first attempt to prompt an LLM with a causal component for the sake of BP explainability. 
Correspondingly, our goal is to automate explanations for BPs using LLMs. For this, our research approach is to blend a variety of BP-related views i.e., process, causal, and \chadded[id=add]{eXplainable AI}\chdeleted[id=del]{XAI.}\chadded[id=add]{, employing state-of-the-art techniques to ensure the correctness of these ingredients}. The combination of these views forms the fuel for the automatic generation of narratives that serve as input for prompt engineering of LLMs to achieve better \chadded[id=add]{perceived} process outcome explanations as illustrated in Figure~\ref{fig:general-approach}. 

The paper's contribution is in providing a tool for integration with LLMs to automate such explanations articulation in BPs, examining how such explanations may be shaped by different inputs to the LLM and a methodological evaluation of the \chadded[id=add]{perceived} quality of such explanations \chadded[id=add]{yielding a designated scale}.

Our paper is structured as follows: 
We begin by reviewing some fundamental concepts in Section~\ref{sec:background} and corresponding research hypotheses in Section~\ref{sec:question-hyp}. This is followed by presenting our designed and developed tooling in Section~\ref{sec:sax-tooling}, and an illustrative example demonstrating its usage in Section~\ref{sec:illustrative-example}. Subsequently, we describe our method of evaluation in Section~\ref{sec:method}, detail a corresponding experimental design in Section~\ref{sec:experimental-design}, and discuss the results in Section~\ref{sec:results}. We conclude the paper with a review of related work in Section~\ref{sec:related}, and our conclusions in Section~\ref{sec:conclusions}.



\section{Background}
\label{sec:background}

We briefly explain in this section the main concepts related to our work.

\noindent
\textbf{\textit{Large language models (LLMs)}} are deep-learning (DL) models trained on vast amounts of text data to perform various natural language processing tasks. One of their strengths is their ability to perform few-shot and zero-shot learning with prompt-based learning~\cite{LLMprompt2023}.

With the emergence of LLMs, the practice of `prompt engineering' has been explored recently as a new field of research that focuses on designing, refining, and implementing different instructions to guide the output of LLMs in various tasks to achieve a more concise and effective utilization of the LLMs, adapting it to the corresponding skills of the users and the context of the different tasks such as analysis or reasoning~\cite{Mesko2023}. This may include different techniques such as one-shot or few-shot samples, the use of quotes to annotate different parts of a prompt, employing methodological evolution of the interaction via advanced methodologies such as ``Chain of Thought'' (CoT), and configuration of LLM-specific settings like temperature and top-p sampling to control the variability of the output.

In this work, we evaluate the use of LLMs as a means to synthesize a combination of BP-related knowledge ingredients presented in textual form as input to the LLM. Subsequently, we employ the LLM to articulate explanations that correspond to user inquiries about particular situations or `conditions' that may occur during BP executions. 

Business Process Management (BPM) is a managerial effort devoted to ensuring the constant, efficient, and effective movement of a business entity towards its constantly
changing goals. This entails adopting a process view of the business and includes planning, analysis, design, implementation, execution, and control of BPs.
A \textbf{\textit{business process (BP)}} is a collection of tasks that execute in a specific sequence to achieve some business goal~\cite{Weske2019}. The digital footprint that depicts a single execution of a process as a concrete sequence of activities or events is termed a `trace'~\cite{vanderAalst2016ProcessMining}. A multi-set of traces is usually referred to as a trace-log or event-log. 

\textbf{\textit{Process discovery (PD)}} aims at gaining insights into the BPs of organizations by analyzing event data recorded in their information systems for the sake of business improvement~\cite{vanderAalst2016ProcessMining}. PD summarizes an event log $L$ into a graph model $M$ that represents activities and control-flow dependencies~\cite{Leemans2019AutomatedDiscovery}. Typical process mining techniques are inherently associational, approaching process discovery from a time precedence perspective, i.e., they discover ordering constraints among the process’ activities. That is, most PD algorithms construct edges in $M$ that indicate to which subsequent activities process control “flows to”. 

Associative relationships are not enough to develop the causal understanding necessary to inform intervention recommendations as aforementioned. Most statistical learning procedures reflect correlation structures between features instead of their true, inherent, causal structure which is the true goal of explanations~\cite{Zhou2021}.
In order to understand which intervention should be made to improve a process outcome, we first must understand the causal chains that tie the unfolding
of the processes generating our data. Unraveling the causal relationships among the execution of process activities is a crucial element in predicting the consequences of process interventions and making informed decisions regarding process improvements. Our previous work~\cite{Fournier2023TheDependencies} demonstrated that relying only on time precedence between activities in a BP does not necessarily reflect the full cause-effect dependencies among the tasks and a more fundamental analysis of causal relationships among the tasks is required.

\textbf{\textit{Causal discovery (CD)}}  infers causal graphs from data by exploring relationships like $A \xrightarrow{c} B$ where changes in $A$ entail changes in $B$. Causal relationships describe the connection between a cause and its effect, where the cause is an event that contributes to the production of another event, the effect~\cite{Pearl2011Causality:Edition}. Causal discovery is responsible for creating models (causal graphs) that illustrate the causal relationships inherent in the data~\cite{Spirtes2001CausationSearch,Pearl2011Causality:Edition,Qafari2020RootModels}. Identification of causal relationships is key to the ability to reason about the consequences of interventions. One of the fundamental goals of causal analysis is not only to understand exactly what causes a specific effect but rather to be able to conclude if certain interventions account for the formation of certain outcomes, thus, being able to answer questions of the form: Does the execution of a certain activity in a loan approval process entail a delay in the handling of the application? or if the same activity is skipped, may the process duration be shortened? In our work, we focus on causal discovery and adapt for its employment over process execution times, as inherently recorded in process event logs. More specifically, we leverage the Linear Non-Gaussian Acyclic Model (LiNGAM)~\cite{Shimizu2022StatisticalApproach} for CD as in~\cite{Fournier2023TheDependencies} to uncover the causal execution dependencies among the events in event logs. Inspired by \cite{vidgof2023large}, which highlights LLMs' ability to provide interpretable explanations, we aim to demonstrate that CD can enhance explanations of process execution outcomes when used as input for LLMs.

The penetration of AI into BPM has vast implications across the various dimensions of BPM~\cite{Dumas2023}, featuring new capabilities to better predict process executions, understand and foresee the implications of changes, and gradually automate the different practices associated with process management. 
However, the efficiency of AI-based solutions is proportionally related to their complexity. As the accuracy and efficiency of machine learning (ML) solutions increase, their complexity increases while reducing human interpretability requiring external explanation frameworks ~\cite{Verma2021,Adadi2018,Meske2022ExplainableOpportunities, Elkhawaga2023}. 

\textbf{\textit{eXplainable AI (XAI)}} is the ability to understand and interpret how AI systems make decisions or arrive at conclusions. Such frameworks are predominately developed for post-hoc interpretations of ML models~\cite{Adadi2018,Meske2022ExplainableOpportunities}. Context-wise, they can be divided into global, local, and hybrid explanations~\cite{Adadi2018,Rehse2019,Guidotti2018}. Global explanations attempt to explain the ML model's internal logic, local explanations try to explain the ML model's prediction for a single input instance, and hybrid approaches vary (e.g., explaining the ML model's internal logic for a subspace of the input space).

In contemporary XAI techniques (e.g., LIME~\cite{Ribeiro2016} or SHAP~\cite{Lundberg2017}), the ML model serves as a surrogate model typically trained using historical process execution logs of the BP. The predicted value for a single instance (process outcome) serves as input for the XAI explainer to produce an explanation. Our work adds to a series of recent efforts~\cite{Amit2022Model,Upadhyay2021ExtendingAutomation} that focus on exploiting XAI frameworks that are compatible with tabular data for the interpretation of BP execution results. We use process logs as the main data input and train surrogate ML models with this data to represent real-world BPs.

However, contemporary techniques are not adequate to produce explanations faithfully and correctly when applied to BPs as they generally fail to\chadded[id=add]{:}
\begin{itemize}
    \item Express the BP model constraints (i.e., the semantics of the process model),
    \item Include the richness of contextual situations that affect process outcomes (additional information that affects the outcome but usually not modeled), 
    \item Reflect the true causal execution dependencies among the activities in the BP, or
    \item Make sense and be interpretable to process users (explanations are usually not given in a human-interpretable form that can ease the understanding by humans).
\end{itemize}


Our goal is to combine PD, CD, and XAI to generate narratives for improved process outcome explanations using LLMs.

To this end, we introduce \textit{\textbf{Situation-Aware eXplainability (SAX)}} as a framework for generating explanations that address the aforementioned shortcomings. 

More specifically, a ``SAX explanation'' is a causally sound explanation that takes into account the process context in which some condition occurred (the ``explanandum''). Causally sound means that the explanation given provides an account of why the condition occurred in a faithful and logical entailment that reflects the genuine chain of BP executions yielding the condition. Context includes knowledge elements that were originally implicit or part of the surrounding system, yet affected the choices that have been made during process execution. 


The SAX framework is realized through a set of services in the SAX4BPM library (Section~\ref{sec:sax-tooling}) that aids with the automatic derivation of SAX explanations leveraging existing LLMs.

\section{\chadded[id=addr2]{Related Work}}
\label{sec:related}

\chadded[id=addr2]{In the following, we discuss how our study relates to and contributes to the literature on 
the intersection between LLMs and BPs for the sake of BP explainability.}

\chadded[id=addr2]{Overall, the use of LLMs for BPs has been researched only very recently, enabled by the availability and accessibility of the GPT-based foundation models. In a broader perspective, BP characteristics present a unique opportunity for the development of a new class of foundation models that utilize the timely sequencing of processes for various process-related tasks such as activity prediction, process optimization, and decision making (e.g.,~\cite{Beheshti2023ProcessGPT:Intelligence,Rizk2024}).
Vigdof et al.~\cite{vidgof2023large} explore the potential and broader application for integrating LLMs along all stages of the BPM life-cycle and propose further broad research directions for the use of LLMs in BPM. While covering process mining, optimization, and decision-making, the paper does not conduct experiments. Grohs et al.~\cite{Grohs2024} demonstrate in several experiments that advanced LLMs such as GPT4.0 are capable of transforming natural language descriptions of smaller processes into declarative and imperative process models and can aid in detecting automate-able tasks from natural language process descriptions. None of the above papers delves into the detailed use of LLMs for explainability in BPs nor examines the consequences of such employment, which is the focus of our study. Berti et al.~\cite{Berti2024} demonstrate the ability to infer process abstractions from event data and subsequently feed these to an LLM along with a query answering process-related questions. While there may be some overlap in eliciting the process view as input for the LLM, our work extends the input to the LLM by incorporating causal and XAI views to derive sound and interpretable explanations.}

\chadded[id=addr2]{Prompt engineering for LLMs that are employed towards various tasks in BPM is also recently surveyed in~\cite{Rizk2024}. While zero-short or few-short learning is reported to improve the performance of the LLMs, other techniques such as modifying its structure, and even subtle reordering of the various input elements, may have destructive effects on the performance of the LLM. These experiences have recently also resulted in a public repository of prompts~\cite{Bach2022}. To mitigate such destructive consequences, Jessen et al.~\cite{Jessen2023} propose a fully automated, structured process for generating prompts for translating natural-language analysis questions into SQL queries over event data. Their approach relies on (iteratively) enriching the prompt with additional information from different perspectives, which then results in a more stable quality of answers (correct SQL queries). While our approach for prompt engineering bears similarities as we also integrate information from multiple perspectives, our focus is on altering the content of the input (semantics) to the LLM rather than reshuffling its structure (syntax). We pursue this to address BP explainability and leverage information in a knowledge graph to automatically generate the prompt.}

\chadded[id=addr2]{A general assumption made by existing studies on LLMs for BPM is that the textual output generated by an LLM is inherently understandable and suitable for the user. While some prior studies do analyze fidelity~\cite{Jessen2023}, this study is the first to also investigate the interpretability of LLM output by users and other background factors that influence these. Notably, we establish that users perceive a trade-off between fidelity and interpretability.}

\chadded[id=addr2]{Several works studied the application of explainability techniques for process prediction and prescription. For example, Shapley Values~\cite{shapley:book1952} have been used to explain which features influence the prediction made by an ML model learned from an event log~\cite{GalantiCLCN2}. Shapley values are also used as a feature for generating explanations for the recommendations produced by a prescriptive model~\cite{Padella2022}. Several other works also explore (combinations of) other ``explainability'' techniques for process prediction and recommendation, e.g.,~\cite{El-Khawaga2022,Stevens2021,Velmurugan2021,Wickramanayake2023}. 
Generally, this line of research focuses on explaining a predictive or prescriptive model of a process, i.e., it aims to explain how an approximation of the observed event log relates input to output features. We extend the conventional applicability of XAI techniques to BP explainability to also include other perspectives such as causal and process views which are inferred directly from the process event log.}

\chadded[id=addr2]{Further, all studies have in common that they generate explanations in the form of charts that visualize the influence of input on output features. A user study confirmed that the generated visualizations ``are generally comprehensible to correctly carry out analysis’ tasks'' for prediction~\cite{Galanti2023AnAnalytics}. The study also observed that accuracy in understanding influences on outcomes and the overall quality of the approach (esp. on difficult tasks) could improve when analysts are better aided in understanding how different influencing factors integrate towards the final effect.  Another user study~\cite{Rizzi2022} showed that also experts undergo a learning curve when interpreting the plots and require additional explanations for answering more difficult tasks. Our study contributes to integrating various features for explanations and complements these prior studies by analyzing the factors that influence the understandability of explanations in natural language.}

\chadded[id=addr2]{A novel component in this work is the incorporation of the causal execution view as a basis for generating high-fidelity explanations. With regards to the applicability of causal inference and discovery to BPs, recently, Dasht Bozorgi et al. demonstrated how to apply statistical causality analysis techniques to identify cause-effect relations in BPs from event logs~\cite{Bozorgi2020}. These causality analysis techniques can be used to improve control policies using reinforcement learning~\cite{Bozorgi2021PrescriptiveReduction} or to improve prescriptive process monitoring by estimating the causal effect of an intervention on a performance metric~\cite{DashtBozorgi2023PrescriptiveEstimation}. 
While Dasht Bozorgi et al. focus on identifying individual rules of cause-effect of an action utilizing the Conditional Average Treatment Effect (CATE) technique~\cite{Kunzel2019MetalearnersLearning}, our work employs a novel approach that utilizes the timing of the activities for the discovery of causal execution interdependency between process activities, and use this view as an additional input to the LLM.}

\section{Hypotheses}
\chadded[id=add]{
\label{sec:question-hyp}
With our intention to employ LLMs as instrumentation for the generation of \chdeleted[id=del]{process-related} SAX explanations, we have set up our goal to tackle the following research question:\\
\noindent
\textbf{Research Question:} \textit{How does having informed knowledge of business processes influence the perceived quality of explanations generated by LLMs regarding business process conditions?}\\
}
\chadded[id=add]{Where a \textit{condition} in a BP relates to any possible state or outcome that may be brought about during the execution of the BP. In an effort to empirically address this question, we developed the underlying software artifact (see section~\ref{sec:sax-tooling}) as a means to manipulate knowledge about BPs to tackle the first part of the question. Respectively, to tackle its second part, we designed an experiment to assess the influence of the manipulation on the perceived quality of the explanations (see section~\ref{sec:method}). For its operationalization, we reformulated the `perceived quality of LLM-generated explanations' into its two main constructs (i.e., dependent variables), as identified in prior literature: namely, `fidelity' and `interpretability'. Each of the two constructs was further instantiated by the formation of the underlying measurement dimensions (i.e., as formative constructs~\cite{Freeze2007}) as further elaborated in the scale development section~\ref{sec:scale-development}. From this, we derived two high-level hypotheses as follows.}

\newtheorem{hypothesis}{Hypothesis}
\chadded[id=add]{
\begin{hypothesis}[H\ref{hyp:first}] 
\label{hyp:first}
Explanations generated by LLMs informed by knowledge about business processes will be perceived as having higher \textbf{fidelity} compared to explanations generated by uninformed LLMs.
\end{hypothesis}}

\chadded[id=add]{
\begin{hypothesis}[H\ref{hyp:second}] 
\label{hyp:second}
Explanations generated by LLMs informed by knowledge about BPs will be perceived as having higher \textbf{interpretability} compared to explanations generated by uninformed LLMs.
\end{hypothesis}}

\section{SAX tooling (towards a plugin for LLM)}
\label{sec:sax-tooling}

\chadded[id=add]{From a design science perspective~\cite{Johannesson2014}, our main developed artifact is}
the SAX4BPM library. \chadded[id=add]{The purpose of its development is to include the functionality needed for the elicitation of three knowledge ingredients from a given process event log: causal, process, and XAI. Different combinations of these are formed as prompt input for an LLM, preceding query answering about various process conditions.}

\subsection{SAX4BPM Library Design}

\chadded[id=add]{The aforementioned research question unfolds from the problem of using LLMs out-of-the-box, without having such interactions ``guard railed'' by supplementing their input with adequate knowledge about the BPs inquired about. 
We acknowledge that the derivation of sound explanations is influenced by both contextual knowledge and the ability to reason correctly over this knowledge. Our focus in this work is on manipulating the former while neutralizing the effect of the latter by conducting a comparative analysis of the output, fixating the use of the same LLM across all tests, and embedding ground truth in the experimental protocol. Such a comparative analysis also eliminates the need to consider the a priori performance of the LLM, making fine-tuning and augmentation techniques orthogonal to our examination.}

\chadded[id=add]{In this section, we elaborate on the characteristics of the developed library, highlighting its requirements as a basis for the realization of its various functionalities. Overall, the library is developed as open source, adhering to an incremental development process, and utilizing open and real-world datasets for its applicability testing~\cite{Fournier2023TheDependencies}}. 

\chadded[id=add]{
We have identified four cornerstone requirements, three entailed the need to inform LLM prompts with process-related knowledge, and a fourth related to their blending in the context of invoking the LLM for query answering:}

\begin{itemize}
    \item \chadded[id=add]{Process-related knowledge - Capturing the time precedence relationships among the activities in the BP. This is required to ensure that explanations about the process respect the correct flow of execution among the activities~\cite{Amit2022Model}.}
    \item \chadded[id=add]{Causal-related knowledge - Denoting the causal dependencies among the execution timing of the activities in the BP. This is required to ensure that explanations about the process respect the causal dependencies among the activities in the process~\cite{Fournier2023TheDependencies}.}
    \item \chadded[id=add]{XAI-related knowledge - Depicting the relative importance of the various activity features concerning some process execution outcome. This is required to ensure that explanations about the process consider the degree of effect of the various activity features on the inquired outcome~\cite{Amit2022Model}.}
    \item \chadded[id=add]{A synthesized prompt - The construction and execution of a prompt that contains a blending of the above knowledge ingredients jointly with the user's query for explanation derivation by an LLM.}
\end{itemize}

\chadded[id=add]{The first three requirements are materialized via a designated service that yields a corresponding view based on the process event log as the input (see section~\ref{sec:knowledge-ingredients}). The fourth requirement is materialized via a designated service as explained in section~\ref{sec:knowledge-synthesize}. To ensure adequacy in the implementation of the functionalities and correctness of its output, our developed tooling leverages state-of-the-art algorithms in all three aspects: process mining, XAI, and causal discovery as described below.}

The developed library includes a set of services and capabilities to support the generation of SAX explanations. The library is implemented using Python 3.9 programming language and is released as open-source, accessible at: \url{https://github.com/IBM/Sax4bpm}.

\subsection{Knowledge graphs}

All data persistence for the SAX4BPM library is realized in the form of a knowledge graph (KG)~\cite{Ehrlinger2016}. 

We use a \emph{Labeled Property Graph} (LPG)\cite{Bonifati2018} as a data model, as it allows to store process-related data more compactly than RDF-based knowledge graphs~\cite{DBLP:journals/jodsn/EsserF21}. An LPG $G$ is a directed multi-graph where each node (modeling an entity or concept) is typed by one or more \emph{labels}, and each directed edge (modeling a relationship between entities/concepts) is typed by exactly one label. Each node and relationship has \emph{properties} in the form of attribute-value pairs.

\begin{figure}[ht]
    \centering
    \includegraphics[width=\textwidth]{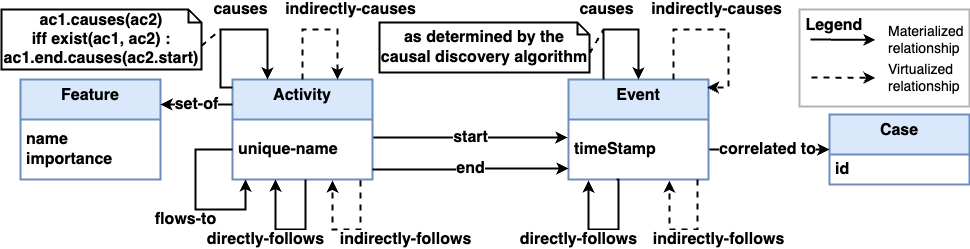}
    \caption{\chadded[id=add]{Knowledge graph schema}}
    \label{fig:graph-schema}
\end{figure}

We use a multi-layered LPG stored in a Neo4j database as infrastructure for our KG and use Neo4j's query language Cypher for \emph{knowledge inference}. We use the schema of~\cite{DBLP:journals/jodsn/EsserF21} as the ``base layer'' for our KG as shown in the graph schema in Fig.~\ref{fig:graph-schema}. Each \emph{Event} node has a \emph{timestamp}, 
and is correlated to one \emph{case}. \emph{Inference rules} over events and cases infer the \emph{directly-follows} relation that describes the temporal order of all events correlated to the same case. These concepts allow modeling any event log in a KG~\cite{DBLP:journals/jodsn/EsserF21}.

The graph schema can be extended with additional ``layers'' of nodes and relations for \emph{integrating} a variety of views on the process in the KG. For example, the schema can be extended with an \emph{Entity} node to enable a multi-dimensional or ``object-centric'' view of the process~\cite{DBLP:books/sp/22/Fahland22}. In our case, we extended the schema 
with the \emph{activity} and \emph{feature} nodes (the former holding a \emph{set-of} the latter), and the \emph{flows-to} and \emph{causes} relations for the following three activity-level inferred views:
\begin{sloppypar}
\begin{itemize}[itemsep=0em,topsep=0pt,leftmargin=*]
    \item \emph{Process view} - we infer \emph{activities} and \emph{directly-follows} ordering 
    of activities by aggregation from events~\cite{DBLP:journals/jodsn/EsserF21}. Accordingly, we infer the \emph{flows-to} relation through process discovery, e.g.,~\cite{DBLP:journals/kais/AugustoCDRP19}. \chadded[id=add]{Respectively, \emph{indirectly-follows} is inferred as a transitive closure of \emph{directly-follows}.}
    \item \emph{Causal view} - we infer the \emph{causes} relationship as a \emph{causal-execution-dependency} as in~\cite{Fournier2023TheDependencies}. Accordingly, we infer the \emph{causes} relation among the activities as annotated. \chadded[id=add]{Respectively, \emph{indirectly-causes} is inferred as a transitive closure of \emph{causes}.}
    \item \emph{XAI view} - we capture the set of \textit{features} and their corresponding \textit{importance} values in the context of each activity.
\end{itemize}
\end{sloppypar}

\subsection{Elicitation of process knowledge ingredients}
\label{sec:knowledge-ingredients}
\begin{figure}[ht]
    \centering
    \includegraphics[width=\textwidth]{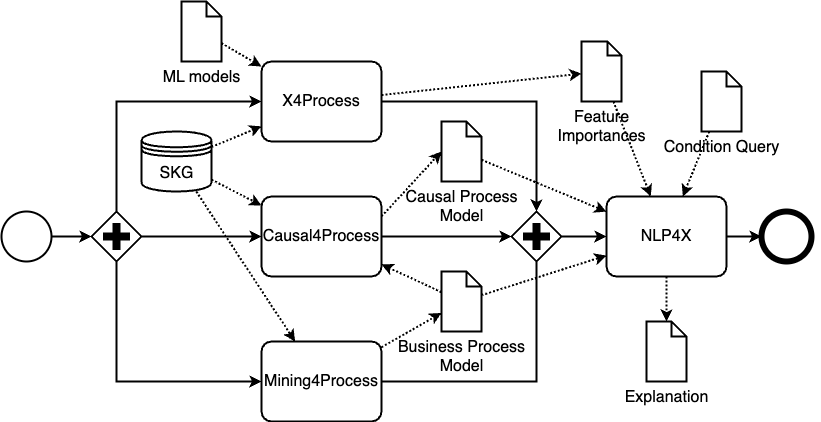}
    \caption{SAX4BPM services invocation}
    \label{fig:services-diagram}
\end{figure}

The library includes a set of functionalities or services for the elicitation of the process knowledge ingredients applied in our study, namely: process, causal, and XAI. Figure~\ref{fig:services-diagram} depicts the flow of the SAX4BPM services yielding an explanation.


\texttt{\textbf{Mining4Process}} – The goal of this service is to generate the process model knowledge ingredient. This service fetches the process event log data from the KG and forwards it for the invocation of an external process discovery algorithm (e.g., heuristic miner~\cite{DBLP:conf/bpm/MannhardtLR17}). \chadded[id=add]{Our current implementation of this service leverages PM4PY~\cite{Berti2023PM4Py:Python}}. The result process model (a graph) is stored back in the KG and also exported in JavaScript Object Notation (JSON) format as input to the LLM.

\texttt{\textbf{Causal4Process}} – 
The goal of this service is to generate the causal process model knowledge ingredient. This service fetches the process event log data and the process model from the KG and employs a novel algorithmic approach~\cite{Fournier2023TheDependencies} to elicit the causal process model. This algorithm leverages an existing causal discovery algorithm (i.e., LiNGAM), and complements it with the needed adjustments for process execution times. The result graph model is stored back in the KG and exported in JSON format as input to the LLM.

\texttt{\textbf{ContextEnrichment}} – The goal of this service is to enrich the event log with any contextual information associated with the process and the environment in which it executes. As an example of such enrichment, our experimentation with complex event processing~\cite{Amit2023} shows that temporal contextual information can be leveraged to improve the adequacy of explanations given for process execution instances. The effect of the enrichment manifests itself not just in properly adjusting the importance of features that correctly correspond to the outcome, but also in promoting the accuracy of the surrogate ML model. 
This service fetches the process event log from the KG and applies a set of situational rules to enrich the event log. The result is an enriched event log which is stored back in the KG.

\texttt{\textbf{X4Process}} – The goal of this service is twofold. First, to generate and determine importance ranking over the set of features that are used in the process to predict the condition of interest while adhering to the constraints of the process model. Second, it uses the \texttt{ContextEnrichment} service to enrich the event log with broader, context-relevant attributes, about the same condition. This service invokes the \texttt{ContextEnrichment} service from the library and subsequently 
fetches the enriched event log and the process model (including constraints and flow control logic) from the KG, both used to constrain the search space of an invoked XAI service (e.g., SHAP or LIME) to ensure conformance to process~\cite{Amit2022Model}. The result is a feature importance vector, segmented according to the activities in the process, that is stored back in the KG and exported in JSON format as input to the LLM.

\subsection{Synthesize of knowledge elements for LLM input}
\label{sec:knowledge-synthesize}
We consider different possible combinations of knowledge ingredients as input for an LLM to facilitate the derivation of explanations about concrete process conditions.

\texttt{\textbf{NLP4X}} – The goal of this service is to interweave the various knowledge ingredients and act as a facade for the interaction with the user, prompting this composition to an LLM. 
This service receives as input the query (the condition the user seeks an explanation for) and the selection of knowledge ingredients. 
The core functionality of the service is the synthesis of the input ingredients by invoking the corresponding SAX4BPM services and their textual streamlining via the LLM prompt toward the eventual elicitation of the explanation narrative.
While the current implementation is hard-coded with predetermined leading phrases to facilitate the interaction, future work aims to automate this step by applying dynamic templating. Our library currently applies \chadded[id=add]{GPT4.0}\chdeleted[id=del]{ChatGPT}\footnote{\url{https://chat.openai.com}}  from OpenAI as the LLM.

\subsection{\chadded[id=add]{Library Architecture}}

\begin{figure}[ht]
    \centering
    \includegraphics[width=1\linewidth]{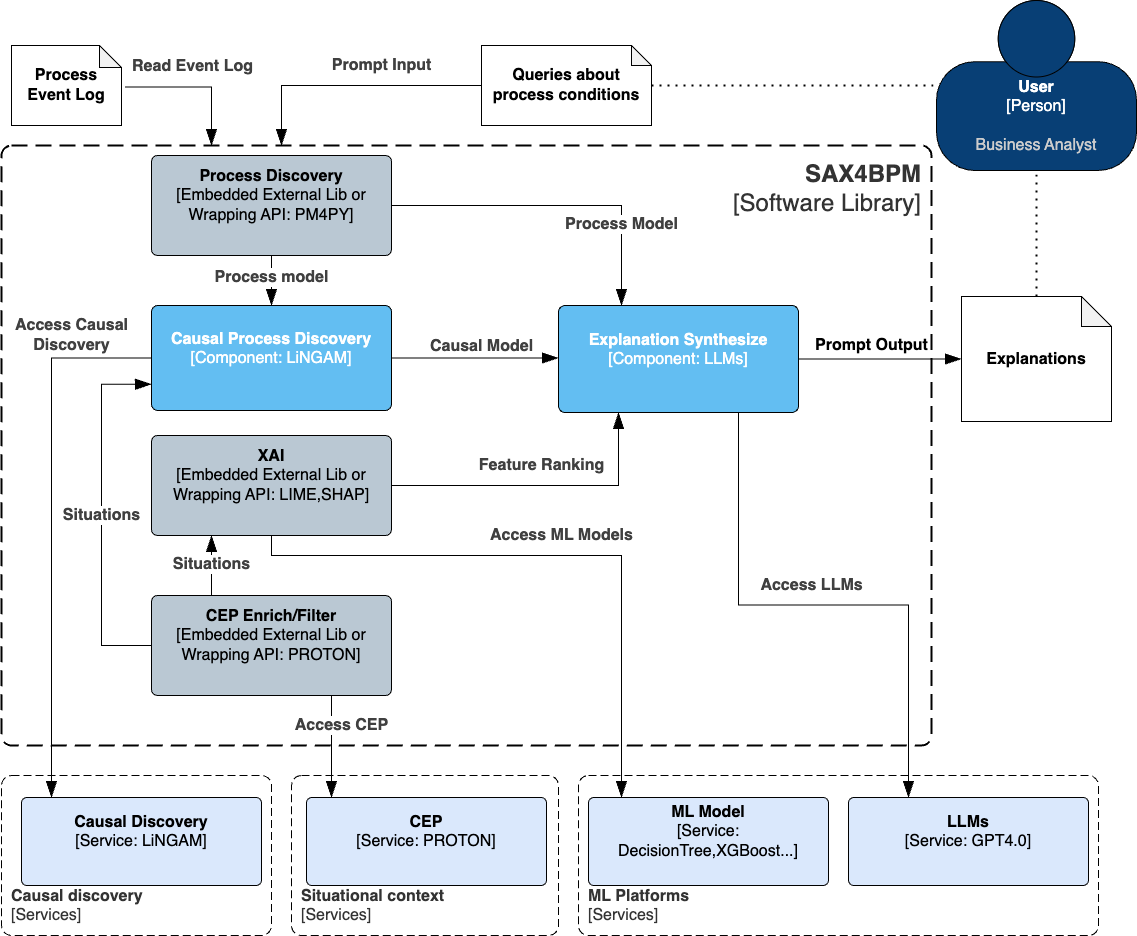}
    \caption{\chadded[id=add]{SAX4BPM Architecture}}
    \label{fig:sax4bpm-architecture}
\end{figure}

\chadded[id=add]{Figure~\ref{fig:sax4bpm-architecture} depicts the different modules comprising the SAX4BPM library, respective to the services aforementioned. 
In general, we classify the modules into modules that leverage existing and external services for the sake of SAX explanations and ones that are developed as part of the library. The first category (grey color) includes the Process Discovery, XAI, and CEP Enrich/Filter modules. The second category (in blue) includes the Causal Process Discovery and the Explanation Synthesize modules. The third category (light blue) relates to external services or packages our library utilizes. We briefly describe each of the internal modules next.}

\noindent
\texttt{Process Discovery} – This module embeds the \texttt{Mining4Process} service functionality to generate the process model out of the event logs.

\noindent
\texttt{Causal process discovery} – This module embeds the \texttt{Causal4Process} service functionality to generate a causal graph based on the execution times of the activities.

\noindent
\texttt{XAI} – This module embeds the \texttt{X4Process} service functionality for the sake of capturing preference importance ranking in the explanation.

\noindent
\texttt{CEP Enrich/Filter} – This module embeds the \texttt{ContextEnrichment} service functionality 
to enrich the event log with additional contextual information or to filter out non-relevant events.

\noindent
\texttt{Explanation Synthesize} - This module embeds the \texttt{NLP4X} service functionality for synthesizing the output from other models and applying the user's query via LLM prompting.

\section{Illustrative example: Parking fines}
\label{sec:illustrative-example}

To illustrate the application of the SAX4BPM services, we generated data on a BP of parking fines. 

\begin{figure}[ht]
    \centering
    \includegraphics[width=\textwidth]{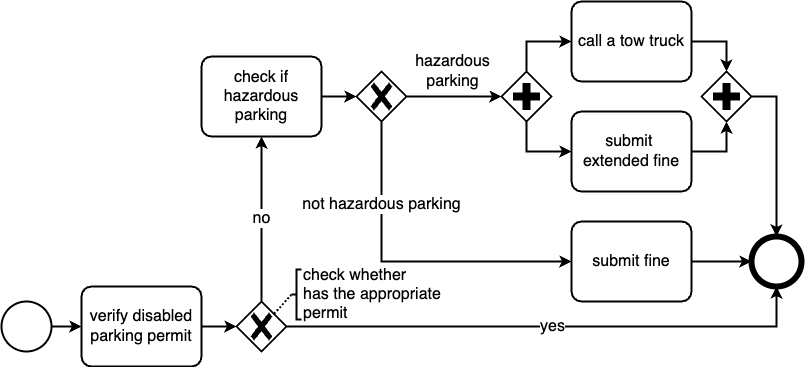}
    \caption{Parking fines scenario}
    \label{fig:parking-scenario}
\end{figure}

In this process (see Figure~\ref{fig:parking-scenario}), a parking ticket is given when a vehicle is parked in a prohibited lot and does not possess a disabled permit. In this case, two types of fines can be given depending on whether the parking place is hazardous (e.g., the vehicle is parked on a sidewalk or a crosswalk) or not. In the case of a hazardous place, an extended fine is issued, and a tow truck is called. Note that the arrival time of the tow truck is always later than the time taking to submit the extended fine.

We generated the dataset for this example using BIMP open-source log simulation tool\footnote{\url{https://bimp.cs.ut.ee/simulator/}} and stored it in the KG.

We demonstrate the capabilities developed in the library through the application of the different services. 


\textbf{The condition for which an explanation is sought:} \textit{Why does the processing of fines for cars that are parked within hazardous locations take so long?}



\subsection{Mining4Process Invocation}
\label{subsec:mining4process}
\begin{figure}[ht]
    \centering
    \includegraphics[width=0.5\textwidth]{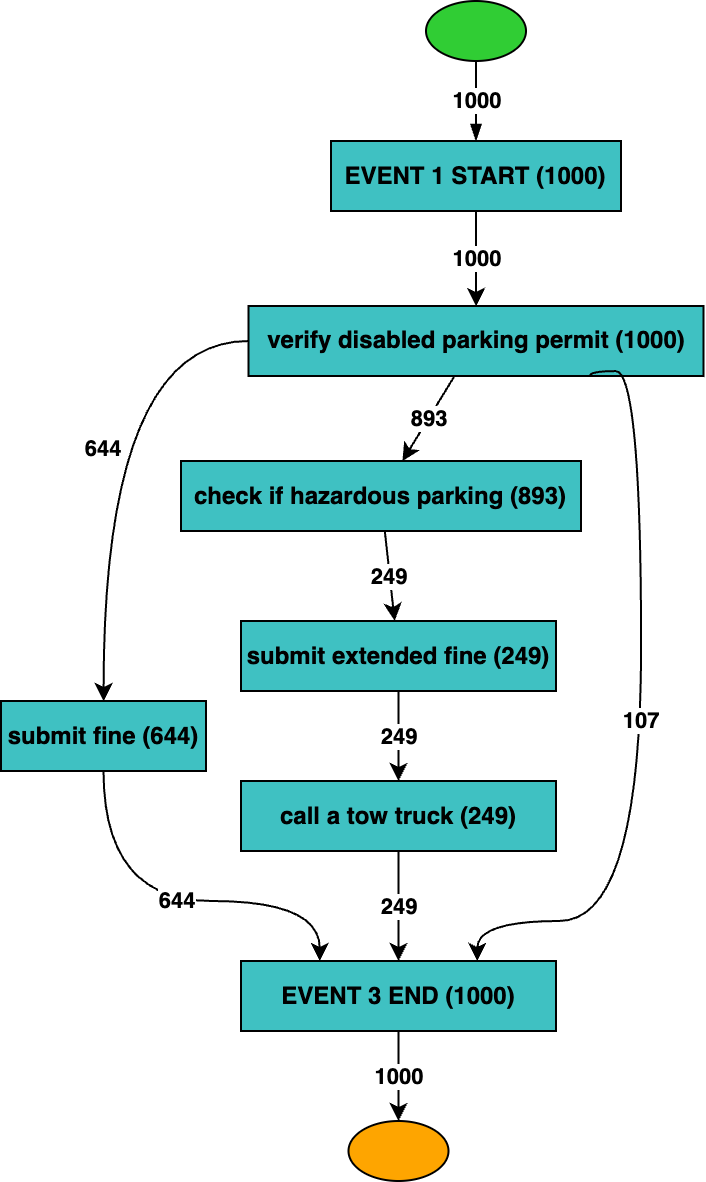}
    \caption{Process model discovery for the parking fines scenario}
    \label{fig:discovered-process}
\end{figure}

The next step is invoking the \texttt{Mining4Process} service to discover a process model from the event log stored in the KG. Figure~\ref{fig:discovered-process} shows the process model discovered in our scenario of parking fines. In the context of this concrete example, we further zoomed in on the process model on the branch corresponding to the issuance of fines for cars parked at hazardous locations (with 249 cases). The generated \chadded[id=add]{text}\chdeleted[id=del]{JSON} that serves as an input for the LLM \chadded[id=add]{prompt adheres to the template:\\ \texttt{\{(a:activity.unique-name,b:activity.unique-name): frequency(a.flows-to(b))\}},\\ and is fetched \chadded[id=add]{in JSON format} via an SKG query} \chadded[id=add]{yielding the following:}\chdeleted[id=del]{is as follows:}

\begin{small}
\begin{spverbatim}
{('EVENT 1 START', 'verify disabled parking permit'): 1000,
('verify disabled parking permit', 
            'check if hazardous parking'): 893,
('check if hazardous parking', 'submit fine'): 644,
('submit fine', 'EVENT 3 END'): 644, 
('check if hazardous parking', 'submit extended fine'): 249, 
('submit extended fine', 'call a tow truck'): 249, 
('call a tow truck', 'EVENT 3 END'): 249, 
('verify disabled parking permit', 'EVENT 3 END'): 107}
\end{spverbatim}
\end{small}

\subsection{Causal4Process Invocation}
\label{subsec:causal4process}

\begin{figure}[ht]
    \centering
    \includegraphics[width=0.8\textwidth]{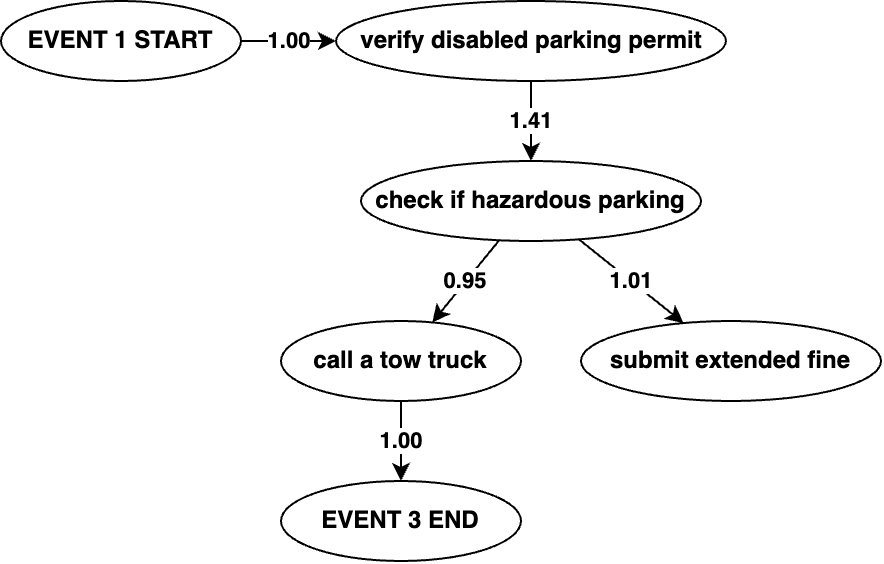}
    \caption{Causal graph discovery for the parking fines scenario}
    \label{fig:variant-causal}
\end{figure}

The next step is invoking the \texttt{Causal4Process} service to discover the causal model. Note that the two consecutive tasks - ``submit extended fine'' and ``call a tow truck'' - in the discovered process model are not causally dependent in the causal model and can be carried out concurrently. Figure~\ref{fig:variant-causal} shows the causal model discovered in our scenario of parking fines. The generated \chadded[id=add]{text}\chdeleted[id=del]{JSON} that serves as an input for the LLM \chadded[id=add]{prompt adheres to the template:\\ \texttt{\{`Cause':(a:activity.unique-name),`Effect':(b:activity.unique-name),\\`Coefficient': coefficient(a.causes(b))\}},\\ and is fetched in JSON format via an SKG query yielding the following:}\chdeleted[id=del]{is as follows:}

\begin{small}
\begin{spverbatim}
{`Cause':`EVENT 1 START', 
        `Effect':`verify disabled parking permit', 
        `Coefficient`:`1.00'}
{`Cause':`verify disabled parking permit', 
        `Effect':`check if hazardous parking',
        `Coefficient':`1.41432045'},
{`Cause':`check if hazardous parking', 
        `Effect':`call a tow truck',
        `Coefficient':`0.95186106'},
{`Cause':`check if hazardous parking', 
        `Effect':`submit extended fine',
        `Coefficient':`1.00753386'},
{`Cause':`call a tow truck', 
        `Effect':`EVENT 3 END', 
        `Coefficient':`1.00'}
\end{spverbatim}
\end{small}

\subsection{X4Process Invocation}
\label{subsec:x4process}

\begin{figure}[ht]
    \centering
    \includegraphics[width=\textwidth]{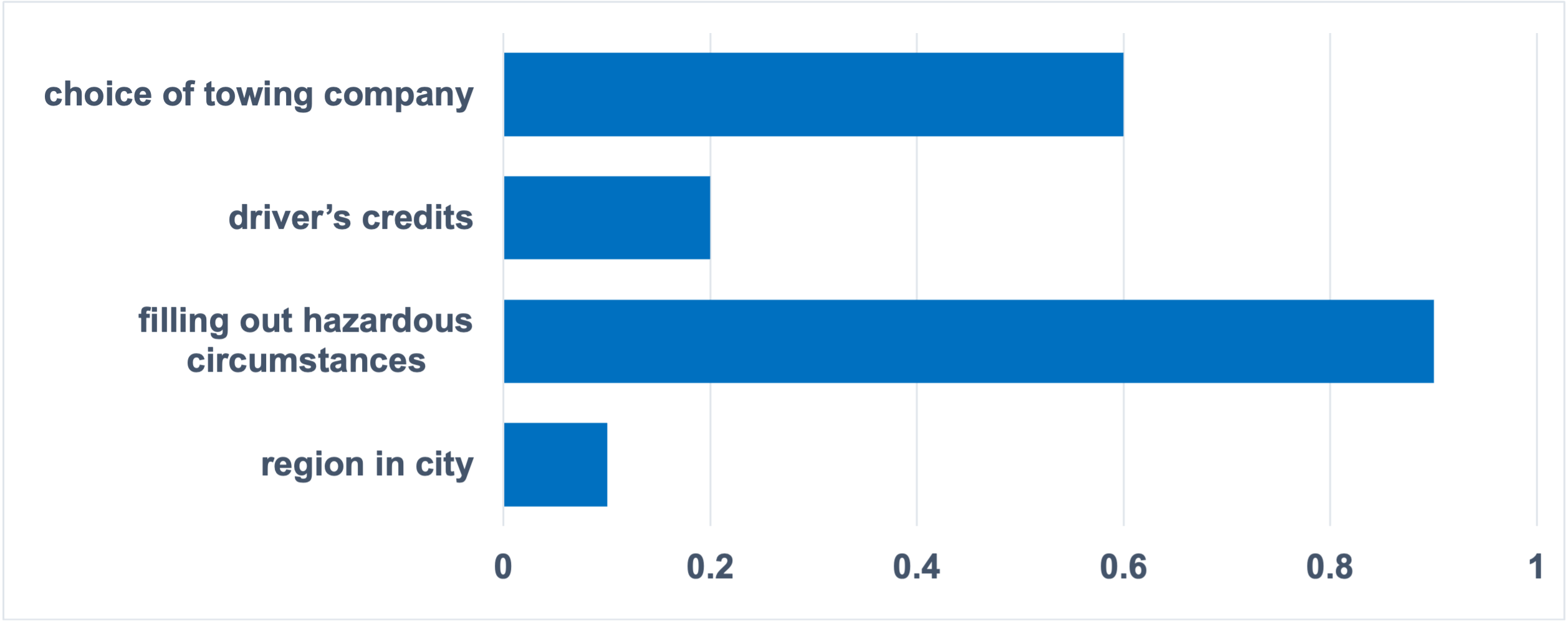}
    \caption{XAI feature importance for the parking fines scenario}
    \label{fig:xai-graph}
\end{figure}

The next step is invoking the \texttt{X4Process} service to elicit the feature importance vector. The context enrichment service enhanced the event log with the following features: ``driver’s credit'', ``filling out hazardous circumstances'', ``region in the city'', and ``choice of towing company''. Figure~\ref{fig:xai-graph} shows the importance of the features discovered in our scenario of parking fines. As can be seen, the ``reason for hazardous location'' is the most important factor causing delays in the process. The generated \chadded[id=add]{text}\chdeleted[id=del]{JSON} that serves as an input for the LLM \chadded[id=add]{prompt adheres to the template:\\ \texttt{\{(a:activity.unique-name), a.setof\{(f:feature.name):(f:feature.importance)\}\}}, and is fetched in JSON format via an SKG query yielding the following:}\chdeleted[id=del]{is as follows:}

\begin{small}
\begin{spverbatim}
{
"check if hazardous parking": 
    {“region in city: 0.1},
"submit extended fine": 
    {"filling out hazardous circumstances": 0.9,
    "driver’s credits": 0.2},
"call a tow truck": 
    {"choice of towing company": 0.6}
}
\end{spverbatim}
\end{small}

\subsection{NLP4X Invocation}

\begin{figure}[ht]
    \centering
    \includegraphics[width=\textwidth]{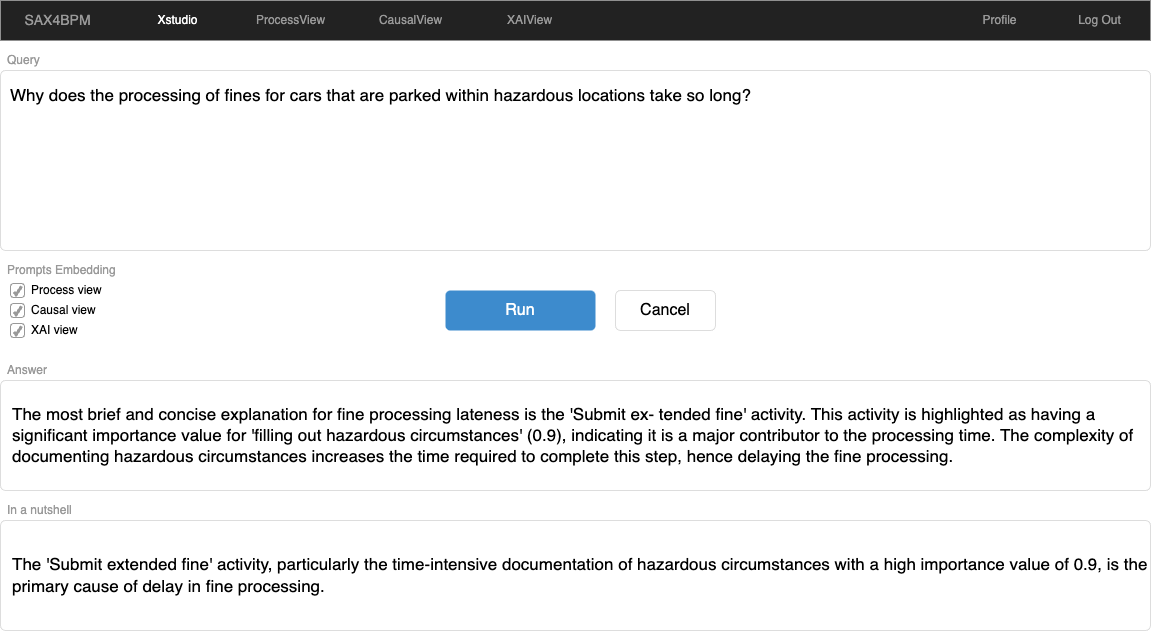}
    \caption{SAX4BPM explanation for the parking fines scenario}
    \vspace{-1em}
    \label{fig:parking-fines-screenshot}
\end{figure}

The last step is invoking the \texttt{NLP4X} service to drive the interaction with the LLM for explanation generation. \chdeleted[id=del]{As per the user's selection of the combination of knowledge ingredients, all three are selected in this example as depicted in Figure~\ref{fig:parking-fines-screenshot}, and the query, the service is invoked.}
\chadded[id=add]{As per the user’s selection of the combination of knowledge ingredients, all three are selected in this example, as depicted in Figure~\ref{fig:parking-fines-screenshot}. Following the query, the service is then invoked.}
In our example, this invocation \chadded[id=add]{generates a blended prompt as in Listing~\ref{lst:blended-prompts}, yielding}\chdeleted[id=del]{yields} the answer as shown in the bottom part of the figure.




\DefineVerbatimEnvironment{MyVerbatim}{Verbatim}{commandchars=\\\{\}, breaklines=true}

\begin{figure}[ht] 
\begin{framed}
\begin{MyVerbatim}
\textbf{PROCESS VIEW:} The following is a JSON list object representing a business process model as determined by the process mining algorithm:
<<--process text part in JSON as in section \ref{subsec:mining4process}-->>
\textbf{CAUSAL VIEW:} <<--causal text part in JSON as in section \ref{subsec:causal4process}-->>
\textbf{XAI VIEW:} In the following JSON object, each element name matches the name of a process activity and contains associated features that may explain process outcomes along with their importance values.
<<--XAI text part in JSON as in section \ref{subsec:x4process}-->>
The above text includes three perspectives about a business process, consisting of a process view, a causal view, and an XAI view. 
\textbf{QUESTION:} Can you give the briefest and most concise explanation to <<--user's query in the UI-->>, considering the views above: [process view], [causal view], and [XAI view]?
\end{MyVerbatim}
\end{framed}
\vspace{-1.3em}
\captionsetup{type=lstlisting}
\caption{\chadded[id=add]{A blended prompt example for all three perspectives}}
\label{lst:blended-prompts}
\end{figure}




\noindent
\chdeleted[id=del]{\textbf{5. Hypotheses\\}}
\chdeleted[id=del]{With our intention to employ LLMs as instrumentation for the generation of process-related explanations, we have set up our goal to tackle the following research question:\\}

\noindent
\chdeleted[id=del]{
\textbf{Research Question:} \textit{How does having informed knowledge of business processes influence the perceived quality of explanations generated by LLMs regarding business process conditions?}\\
}

\chdeleted[id=del]{Where a \textit{condition} in a BP relates to any possible state or outcome that may be brought about during the execution of the BP. In an effort to empirically address this question, we designed an experiment that concretely incorporates the various elements of this research question. To formalize it further, we first reformulated the question by operationalizing the `perceived quality of LLM-generated explanations' into its two main constructs (i.e., dependent variables), as identified in prior literature: namely, `fidelity' and `interpretability'. Each of the two constructs was further instantiated by the formation of the underlying measurement dimensions (i.e., as formative constructs~\cite{Freeze2007}) as further elaborated in the scale development section~\ref{sec:scale-development}. From this, we derived two high-level hypotheses as follows.}

\noindent
\chdeleted[id=del]{\textbf{Hypothesis 1} (H1).
Explanations generated by LLMs informed by knowledge about business processes will be perceived as having higher \textbf{fidelity} compared to explanations generated by uninformed LLMs.}

\noindent
\chdeleted[id=del]{\textbf{Hypothesis 2} (H2).
Explanations generated by LLMs informed by knowledge about business processes will be perceived as having higher \textbf{interpretability} compared to explanations generated by uninformed LLMs.}

\section{Method}
\label{sec:method}

We conducted a controlled experiment to assess our hypotheses. The experiment involved between-group manipulations of various knowledge \chadded[id=add]{ingredients}\chdeleted[id=del]{components} related to different BPs. These components were presented as input to an LLM, which was then prompted to generate an explanation — a textual narrative — articulating the reasons for a specific condition occurring during the execution of the BP. The different explanation variants generated under various manipulations were then incorporated into corresponding versions of a questionnaire. This questionnaire included a set of rating scales developed to measure the perceived quality of the explanations. After collecting user ratings, we tested how the manipulations influenced the perceptions across different groups corresponding to the various explanation variants.

We targeted the population of the global Information Technology community that might benefit from recent advances in LLMs to leverage such technology to automate the generation of explanations about situations in their corresponding area of interest. Pragmatically, we exploited an online questionnaire that was disseminated via three channels for the acquisition of responses: graduate students in the computer science faculty at TU/e, employees at IBM research, and the Reddit r/SampleSize online community. Participation was voluntary and anonymous without collecting any personal information. Before their participation, users gave their consent. Ethical approval was given by the Tu/e Ethical Review Board (\#ERB2023MCS42).

The survey included a total of 50 participants. One respondent did not give consent, resulting in 49 completed survey responses. Regarding gender, 28 were men (57.1\%), 19 were women (38.8\%), and 2 chose not to specify their gender (4.1\%). With respect to age, the distribution was as follows: 23 participants (46.9\%) were above the age of 45, 15 participants (30.6\%) were aged between 25 and 35 years, 6 participants (12.2\%) were below 25 years of age, and 5 participants (10.2\%) were between 36 and 45 years old.

\subsection{Scale Development}
\label{sec:scale-development}

To assess the \chadded[id=addr2]{perceived} quality of the explanations generated by the LLM, we first developed scale metrics based on different measurement dimensions for an explanation's quality. For various reasons, to date, there is no agreed framework for the evaluation of explanations. There is no agreed definition of explainability, furthermore, there is no agreement on what an explanation is, and which salient properties should be considered to make it effective and understandable for end users (~\cite{Elkhawaga2023} and~\cite{Vilone2021}). There is a current lack of consensus regarding a common set of properties or agreed metrics that every explainable method should be evaluated against (~\cite{ Elkhawaga2023},~\cite{Sokol2020},~\cite{Lage2018},~\cite{Markus2021},~\cite{Zhou2021}, and~\cite{Carvalho2019}). Lastly, there is no standardized method for evaluation of explanation's quality. 

We aim to assess the quality of SAX explanations given by an LLM \chadded[id=add]{as perceived by users}, therefore the dimensions selected should reflect desiderata applied in the XAI literature tailored to our specific needs. It is noteworthy to stress that we are actually not assessing the quality of explanations given by an XAI method but rather the explanations generated by an LLM when synthesizing inputs from different knowledge ingredients about the process. Our focus was mostly on the intrinsic qualities of explanations and their content rather than features related to the hedonic and exogenous qualities of the explanations related to the user's experience and interaction with the system. We did embed some attitude aspects in the scale, but we considered these more as background factors that could moderate the other perceptions about the quality of the explanations.

We opted to carry out a user study following the common practice (e.g., ~\cite{ Lage2018},~\cite{Markus2021}¸~\cite{Kulesza2013}) with a scale developed to our needs. 

We henceforth detail some of the common dimensions applied in the literature and assess their level of applicability to our case. \chdeleted[id=del]{Whenever it is not explicitly indicated that the specific property or measurement is not taken into account, it is also reflected in our scale.}\chadded[id=add]{Table~\ref{tab:literature-review} lists all key dimensions reviewed, and Table~\ref{tab:concept-definition} summarizes our eventually adapted constructs, dimensions, and their respective definitions.}

\begin{table}[htbp]
\medlarge
\centering
\def\arraystretch{1.3}
\resizebox{\textwidth}{!}{%
\begin{tabular}{|p{0.8cm}|>{\raggedright\arraybackslash}p{3cm}|>{\raggedright}p{3.7cm}|p{20cm}|}
\hline
\textbf{Ref.} & \textbf{Latent construct} & \textbf{Dimension} & \textbf{Definition in reference} \\ \hline
 & \cellcolor[HTML]{C0C0C0} & Robustness & An explanation can withstand small perturbations of the input that do not change the output prediction. Consequently, robustness expresses a low sensitivity of the XAI method to changes in inputs. \\ \cline{3-4} 
 & \cellcolor[HTML]{C0C0C0} & Fidelity & The XAI method should preserve the internal concepts and original behavior of the black box ML model whenever there is a need to mimic that model. \\ \cline{3-4} 
 & \cellcolor[HTML]{C0C0C0} & Causality & The XAI method should maintain causal relationships between inputs and outputs. An ML model is perceived as being more human-like whenever it provides such causal explanations. Therefore, causality is fundamental to achieving a human understanding of the ML model. \\ \cline{3-4} 
 & \cellcolor[HTML]{C0C0C0} & Trust & The extent that the outcomes of an XAI method enable gaining confidence that the ML model acts as intended. \\ \cline{3-4} 
\multirow{-5}{*}{\cite{Elkhawaga2023}} & \multirow{-5}{*}{\cellcolor[HTML]{C0C0C0}} & Fairness & The extent an explanation enables humans to ensure unbiased decisions of the employed ML models. \\ \hline
 & \cellcolor[HTML]{C0C0C0} & Soundness & Measures how truthful an explanation is with respect to the underlying predictive model. \\ \cline{3-4} 
 & \cellcolor[HTML]{C0C0C0} & Completeness & The extent an explanation generalizes well beyond the particular case in which the explanation was produced. \\ \cline{3-4} 
 & \cellcolor[HTML]{C0C0C0} & Contextfullness & The degree the explanation is accompanied by all the necessary conditions for it to hold, critiques (i.e., explanation oddities) and its similarities to other cases. \\ \cline{3-4} 
 & \cellcolor[HTML]{C0C0C0} & Interactiveness & The explanation process should be controllable and interactive. \\ \cline{3-4} 
 & \cellcolor[HTML]{C0C0C0} & Actionable & An explanation that the users can treat as guidelines towards the desired outcome. \\ \cline{3-4} 
 & \cellcolor[HTML]{C0C0C0} & Chronology & The extent an explanation inherently possesses time ordering, typically with users preferring explanations that account for more recent events as their cause, i.e., proximal causes. \\ \cline{3-4} 
 & \cellcolor[HTML]{C0C0C0} & Coherence & The extent an explanation is aligned with background knowledge and beliefs held by the users. \\ \cline{3-4} 
 & \cellcolor[HTML]{C0C0C0} & Novelty & The extent an explanation contains surprising or abnormal characteristics. \\ \cline{3-4} 
 & \cellcolor[HTML]{C0C0C0} & Complexity & The degree an explanation matches the skills and background knowledge of explainees. \\ \cline{3-4} 
 & \cellcolor[HTML]{C0C0C0} & Personalisation & The extent an explainability technique is adjusted to model the users' background knowledge and mental model. \\ \cline{3-4} 
\multirow{-11}{*}{\cite{Sokol2020}} & \multirow{-11}{*}{\cellcolor[HTML]{C0C0C0}} & Parsimony & The extent an explanation is selective and succinct enough to avoid overwhelming the explainee with unnecessary information. \\ \hline
 & \cellcolor[HTML]{C0C0C0} & Causability & The extent an explanation makes the causal relationships between the inputs and the model’s predictions explicit. \\ \cline{3-4} 
 & \cellcolor[HTML]{C0C0C0} & Interpretability & The capacity to provide or bring out the meaning of an abstract concept. \\ \cline{3-4} 
\multirow{-3}{*}{\cite{Vilone2021}} & \multirow{-3}{*}{\cellcolor[HTML]{C0C0C0}} & Understandability & The capacity to make the model understandable by end users. \\ \hline
 &  & Response time for understanding & Time taken to answer questions about the explained situation. \\ \cline{3-4} 
 &  & Accuracy of understanding & Number of correctly answered questions about the explained situation. \\ \cline{3-4} 
\multirow{-3}{*}{\cite{Lage2018}} & \multirow{-3}{*}{Interpretability} & Subjective satisfaction & Self-reported degree of satisfaction. \\ \hline
 &  & \cellcolor[HTML]{C0C0C0} & How understandable an explanation is for humans. \\ \cline{3-4} 
 &  & Clarity & The degree an explanation is unambiguous. \\ \cline{3-4} 
\multirow{-3}{*}{\begin{tabular}[c]{@{}l@{}}\cite{Markus2021} \\ \cite{Gilpin2018}\end{tabular}} & \multirow{-3}{*}{Interpretability} & Parsimony & The extent an explanation is not too complex, i.e., presented in a compact form. \\ \hline
 &  & \cellcolor[HTML]{C0C0C0} & How accurately an explanation describes model behavior, i.e., how faithful an explanation is to the model. \\ \cline{3-4} 
 &  & Completeness & The extent an explanation describes the entire dynamic of the ML model. \\ \cline{3-4} 
\multirow{-3}{*}{\begin{tabular}[c]{@{}l@{}}\cite{Markus2021}, \\ \cite{Gilpin2018}, \\ \cite{Kulesza2013}\end{tabular}} & \multirow{-3}{*}{Fidelity} & Soundness & The degree an explanation is correct, i.e., truthful to the model. \\ \hline
 & \cellcolor[HTML]{C0C0C0} & Causability & The measurable extent to which an explanation achieves a specified level of causal understanding. \\ \cline{2-4} 
 &  & Clarity & The extent an explanation is unambiguous. \\ \cline{3-4} 
 &  & Parsimony & The extent an explanation is presented in a simple and compact form. \\ \cline{3-4} 
\multirow{-4}{*}{\cite{Zhou2021}} & \multirow{-3}{*}{Interpretability} & Broadness & How generally applicable the explanation is. \\ \hline
 & \cellcolor[HTML]{C0C0C0} & Satisfaction & The degree to which users feel that they sufficiently understand the AI system or process being explained to them. \\ \cline{3-4} 
\multirow{-2}{*}{\cite{Hoffman2023}} & \multirow{-2}{*}{\cellcolor[HTML]{C0C0C0}} & "Goodness" & A set of yes/no questions reflecting explanation correctness, comprehensiveness, coherence, and usefulness. \\ \hline
\cite{Holzinger2020} & \cellcolor[HTML]{C0C0C0} & Causability & The extent to which an explanation of a statement to a user achieves a specified level of causal understanding with effectiveness, efficiency, and satisfaction in a specified context of use. \\ \hline
\end{tabular}%
}
\caption{\chadded[id=add]{Key constructs and dimensions for quality of explanations reviewed in prior literature}}
\label{tab:literature-review}
\end{table}

\chadded[id=add]{With regards to}\chdeleted[id=del]{In}~\cite{Elkhawaga2023}, \chdeleted[id=del]{the authors relate to the following characteristics of explanations:}\chadded[id=add]{we considered its concept of robustness in its broader sense, having embedded it in our completeness dimension. We also abide by its fundamental view of the notion of causality being essential for human understanding, according to which an ML model is perceived as being more human-like whenever it provides such causal explanations. Concerning fairness. we assumed that the inputs generated by the knowledge ingredients possess fairness, therefore if the resulting explanation by the LLM lacks fairness it is related to the soundness of the explanation, which is considered in our scale as described later. }

\chdeleted[id=del]{Trust is the main dimension in~\cite{Lee2004}. The claim is that a trustable technology may be a critical factor in the success of automation and computer technology and that “The operator will tend to trust the automation if its algorithms can be understood and seem capable of achieving the operator’s goals in the current situation”. Furthermore, ~\cite{Lee2004} states that trust can be achieved by making the algorithms of the automation simpler or by revealing their operation more clearly. In our case, this means trust in the tool that generates the explanations, i.e., the LLM.}

\chdeleted[id=del]{In~\cite{Sokol2020} the authors articulate 11 properties of explanations that are important from an explainee’s point of view. Many of these are grounded in social science research.}

\chadded[id=add]{In~\cite{Sokol2020}, the authors identify 11 properties of explanations that are essential from the perspective of the explainee. Many of these properties are rooted in social science research. With respect to its notion of contextfullness, we agree that it helps the user better understand how an explanation can be generalized, thereby allowing the user to assess its soundness and completeness. As shown in section~\ref{sec:experimental-design}, the specific context for each domain use case was given, therefore this dimension was not required. In addition, this metric was also accommodated in our definition of comprehensibility. Likewise, also the notion of an explanation being actionable. Regarding interactiveness, in our empirical experimentation, the interaction was blocked to enable the control for the manipulations of the inputs. Therefore, this property was not taken into consideration in our final scale. The chronology property was inherently represented in the causability dimension in our scale. We also determined coherence and personalization as being tightly related to and included in the comprehensibility measurements in our scale. Adhering to the aspect of an explanation containing surprising and abnormal characteristics, we considered the novelty property as strongly coupled with the perceived curiosity in our scale. Similarly, we associated complexity and parsimony in the dimensions of clarity and compactness, respectively.}

Holzinger et al.~\cite{Holzinger2020} introduce their System Causability Scale to measure the quality of explanations. Causability is also one of the main dimensions in Vilone and Longo~\cite{Vilone2021}. They address the following \chadded[id=add]{three} dimensions: \chadded[id=add]{causability, interpretability, and understandability. As with the aforementioned notion of causality, here also it is stressed that an explanation must make the causal relationships between the input and the model's predictions explicit, especially when these relationships are not evident to the end users. We considered understandability equivalent to our notion of comprehensibility.}


In~\cite{Lage2018}, the prime metrics used to quantify the interpretability of the explanations were the response time for understanding, the accuracy of understanding, and the subjective satisfaction of the users. Both response time and user satisfaction metrics are not essential to the core quality of explanations and therefore were excluded from our scale. Accuracy is covered by comprehensibility and soundness.
In~\cite{Markus2021} and~\cite{Gilpin2018} the authors relate to explainability as the overarching concept, consisting of two main properties: interpretability and fidelity, both necessary to reach explainability. \chdeleted[id=del]{The interpretability of an explanation captures how understandable an explanation is for humans. The fidelity of an explanation expresses how accurately an explanation describes model behavior, i.e., how faithful an explanation is to the model. Furthermore, they consider clarity and parsimony as part of interpretability, and completeness and soundness as part of fidelity.}


Soundness and completeness are also two dimensions of fidelity separately considered in the qualitative study detailed in~\cite{Kulesza2013}.  Here the aim was to investigate how intelligent agents should explain themselves to their users.
The survey paper~\cite{Zhou2021} presents a comprehensive overview of methods proposed in the current literature for the evaluation of ML explanations. One of the main concepts is causability which refers to the measurable extent to which an explanation to a human achieves a specified level of causal understanding, based on causality, which refers to the relationship between cause and effect~\cite{Pearl2011Causality:Edition}. They also refer to interpretability as having the following properties: clarity and parsimony or compactness as presented before, but also broadness. \chdeleted[id=del]{By broadness, the authors mean how generally applicable the explanation is.} We use completeness which embeds broadness and therefore supersedes it.
\chdeleted[id=del]{Curiosity is another dimension worthwhile tackling according to~\cite{Hoffman2023}. The idea is that the seeking of an explanation can be driven by curiosity. The assessment of users’ feelings of curiosity might be informative in the evaluation of XAI systems due to the following:}
\chdeleted[id=del]{Curiosity in our case is translated to the general desire to acquire knowledge about the reason for a certain condition.}

Hoffman et al.~\cite{Hoffman2023} state that in the XAI context, a mental model is a user’s understanding of the AI system and its context. XAI system development requires a method for eliciting, representing, and analyzing users’ mental models. The authors distinguish between two separable things: (1) the intrinsic goodness of explanations and (2) the user’s satisfaction with the explanations. \chdeleted[id=del]{Explanation satisfaction is defined here as the degree to which users feel that they sufficiently understand the AI system or process being explained to them.} Mental model goodness includes correctness (soundness in our set of dimensions), comprehensiveness (completeness in our set of dimensions), coherence (soundness in our set of dimensions), and usefulness (excluded as this is not an intrinsic property of an explanation) while several key attributes of explanation satisfaction include: understandability (comprehensibility in our set of dimensions), feeling of satisfaction (excluded as this is not an intrinsic property of an explanation), sufficiency of detail (included as per the definition of completeness), completeness, usefulness (not relevant), accuracy (included in soundness), and trustworthiness.

\chdeleted[id=del]{Holzinger et al.~\cite{Holzinger2020} introduce their System Causability Scale to measure the quality of explanations. Here causability is defined as the extent to which an explanation of a statement to a user achieves a specified level of causal understanding with effectiveness, efficiency, and satisfaction in a specified context of use.}

\begin{figure}[ht]
    \centering
    \includegraphics[width=1\linewidth]{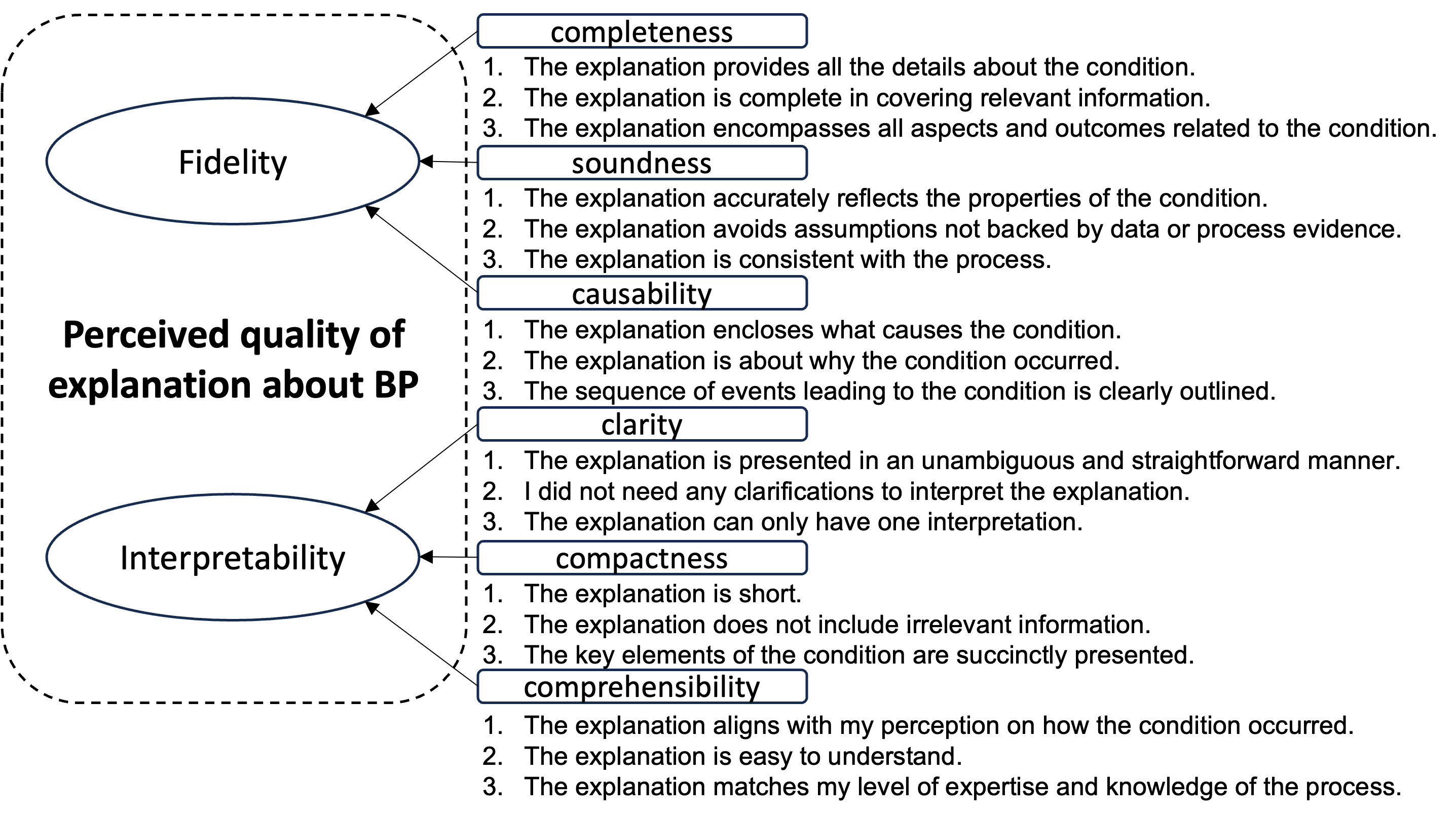}
    \caption{(Perceived) explanation quality – \chadded[id=add]{a total manifestation of 18 questions populated for the measurement dimensions, corresponding to each of the two high-level (latent) constructs}\chdeleted[id=del]{high-level (latent) constructs – as a composition of perceived characteristics}}
    \label{fig:explanation-quality-dimensions}
\end{figure}

Following our survey literature, we adopted fidelity and interpretability as two main perceived latent constructs that subsume some of the measurable dimensions adapted to our user study. As a result, under fidelity, we include completeness, soundness, and causability dimensions. Interpretability includes clarity, compactness, and comprehensibility. We relate causability to fidelity as it refers to the correctness of the underlying model whereas comprehensibility relates to the level of understanding of the explanation and therefore is connected to interpretability. Figure~\ref{fig:explanation-quality-dimensions}  
depicts the six dimensions grouped by the constructs of fidelity and interpretability along with the final questions posed in the survey \chdeleted[id=del]{(explained below)}. 

\chdeleted[id=del]{In addition, for the sake of our study, we also included perceived curiosity (in the problem) and trust (in the LLM) as potential background factors or moderators. The idea behind this selection was that these two dimensions cater more to the interaction formed a-priori to it, rather than reflect an intrinsic quality about the explanation or the content it embeds.}

\chadded[id=addr2]{We acknowledge that the a-priori objective correctness of the explanation may interfere with its eventual perception. To mitigate such influence, this factor could be either measured or controlled for. In our case, we controlled for the correctness of the explanations by having them reviewed by the study researchers (i.e., the authors) who are experts in BPM, and also presented them to the participants as a reference (“ground truth”).}

\begin{figure}[ht]
    \centering
    \includegraphics[width=0.7\linewidth]{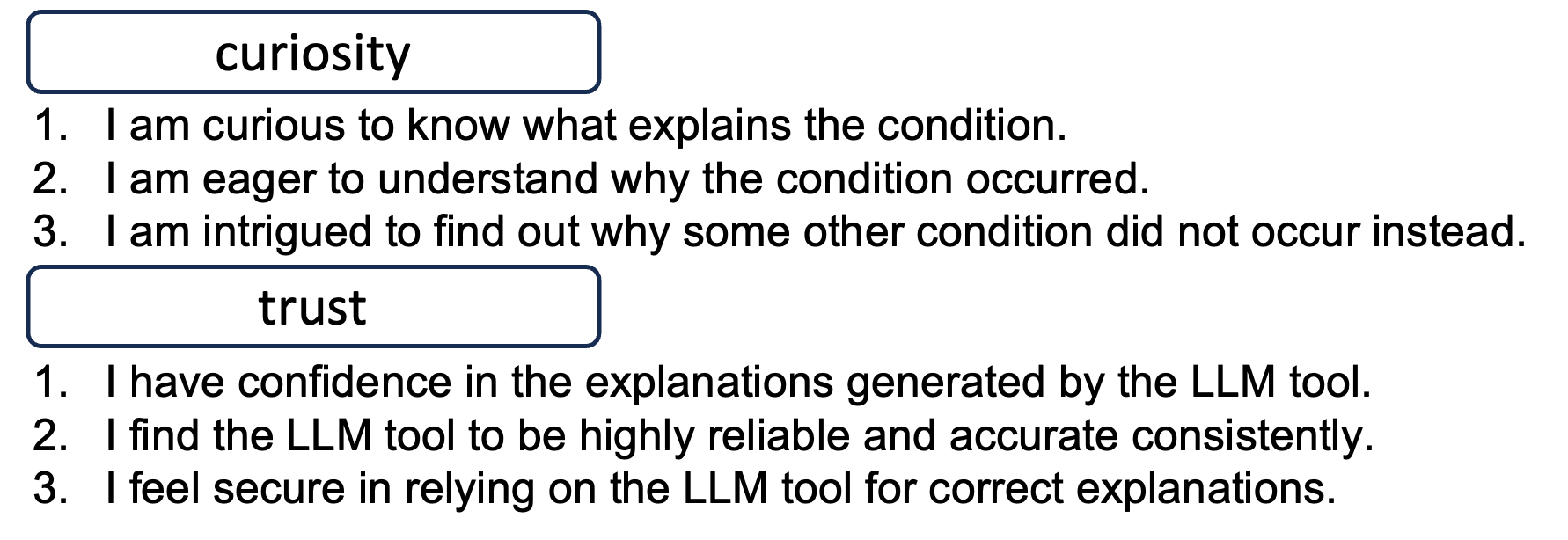}
    \caption{(Perceived) moderating factors (covariates) - \chadded[id=add]{an additional manifestation of 6 questions}}
    \vspace{-1em}
    \label{fig:background-factors}
\end{figure}

Given the nature of our controlled experiment with voluntary participation, we were concerned that the primary effect might be obscured by the participants' interest in the explanations provided to them. \chdeleted[id=del]{Therefore, we also measured and later incorporated `trust' and personal `curiosity' as potential covariates in our analysis.
Figure~\ref{fig:background-factors}. shows these additional two dimensions.}
\chadded[id=add]{We identified curiosity and trust as two relevant constructs related to the manifestation of such an interest. Trust is the main dimension in~\cite{Lee2004}. The claim is that a trustable technology may be a critical factor in the success of automation and computer technology and that ``The operator will tend to trust the automation if its algorithms can be understood and seem capable of achieving the operator's goals in the current situation''. Furthermore, ~\cite{Lee2004} states that trust can be achieved by making the algorithms of the automation simpler or by revealing their operation more clearly. In our case, this means trust in the tool that generates the explanations, i.e., the LLM.}
\chadded[id=add]{Curiosity is another dimension worthwhile tackling according to~\cite{Hoffman2023}. The idea is that the seeking of an explanation can be driven by curiosity. Curiosity in our case is translated to the general desire to acquire knowledge about the reason for a certain condition.}

\chadded[id=add]{Consequently, for the sake of our study, we also included perceived curiosity (in the problem) and trust (in the LLM) as potential background factors (covariates) or moderators. The idea behind this selection was that these two dimensions cater more to the interaction formed a-priori to it, rather than reflect an intrinsic quality about the explanation or the content it embeds. Figure~\ref{fig:background-factors} shows these additional two dimensions.}

Overall, our literature survey concluded having two latent constructs, each operationalized by 3 underlying measurement dimensions (i.e., 6 in total), and two additional background factors. Table~\ref{tab:concept-definition} shows our adapted definitions for all dimensions.

\begin{table}[htb]
\Large
\centering
\def\arraystretch{1.2}
\resizebox{\textwidth}{!}{%
\begin{tabular}{|l|l|l|}
\hline
\textbf{\begin{tabular}[c]{@{}l@{}}Dependent\\ Variables\end{tabular}} & \textbf{\begin{tabular}[c]{@{}l@{}}Measurement\\ Dimensions\end{tabular}} & \textbf{Definition} \\ \hline
Fidelity & \cellcolor[HTML]{C0C0C0} & How faithful an explanation is to the condition explained. \\ \hline
 & Completeness & The extent to which an explanation describes all the information relevant to the condition explained. \\ \hline
 & Soundness & How truthful the explanation is with respect to the condition explained. \\ \hline
 & Causability & The explanation provides the reasons for the occurrence of the condition explained. \\ \hline
Interpretability & \cellcolor[HTML]{C0C0C0} & How understandable an explanation is for humans. \\ \hline
 & Clarity & The extent the explanation is unambiguous. \\ \hline
 & Compactness & The extent the explanation is presented in a simple and compact form. \\ \hline
 & Comprehensibility & The explanation aligns with my understanding of the problem and how to react to it. \\ \hline
\textbf{\begin{tabular}[c]{@{}l@{}}Background\\ Factors\end{tabular}} & \cellcolor[HTML]{C0C0C0} & \cellcolor[HTML]{C0C0C0} \\ \hline
Curiosity & \cellcolor[HTML]{C0C0C0} & The general desire to acquire knowledge about the reason for a certain condition. \\ \hline
Trust & \cellcolor[HTML]{C0C0C0} & An attitude toward the LLM that affects reliance on its explanations. \\ \hline
\end{tabular}%
}
\caption{Definitions for all experimental constructs, measurement dimensions, and background factors}
\vspace{-1em}
\label{tab:concept-definition}
\end{table}


We next populated a preliminary set of questions based on existing scales for each of the 8 measurement dimensions. Before its dissemination, we pursued a card sorting procedure (~\cite{Davis1985,Davis1989,Moore1991}) in order to ensure the validity and reliability of our scale. This was subsequently complemented by a reliability analysis of the responses that we conducted after the survey. 

Card sorting is an iterative scale development technique in which a panel of judges is asked in several rounds to sort a set of scale items into separate categories, based on similarities and differences among them. This technique was previously used in (~\cite{Davis1985,Davis1989}) to assess the coverage of an intended domain of constructs. Later it was further refined~\cite{Moore1991} to generally account for scale reliability (i.e., content and construct validity). 

Prior to conducting the first card sorting round, we populated five items for each category or measurement dimension. We used KardSort\footnote{https://kardsort.com/} as the tool for card sorting.

The set of all 40 items (5 items for each of the 8 dimensions or categories) was jointly considered as the input pool for the first card sorting round. To eliminate ambiguous items, we conducted three separate card sorting rounds with five independent judges in each round. As noted in 
~\cite{Moore1991}, this was done to increase dimension validity and confidence in the developed scale. The judges at each round were presented with all the items and were asked to relate each item to the best-fitted category. 

We used two indicators to monitor validity and the convergence of the sorting procedure: (1) Inter-rater agreement between the judges for each dimension in each round, and (2) Hit ratio (`convergent' validity) - item-wise portion of true item classifications within its `target' dimension (see Table~\ref{tab:card-sorting-result}). 

After each round, we inspected each item's inter-rater and hit ratio values. The inter-rater indicator reflected the cumulative (so far in the rounds) agreement among the judges on a specific item. Items with a hit ratio less than the majority of card assignments were rephrased to better reflect their category. As aforementioned, we performed this adjustment after each round. After the third time, we selected the top 3 items that best fit their category, concluding with a total of 24 items. Table~\ref{tab:card-sorting-result} summarizes the overall procedure whereas Table~\ref{tab:card-sorting-hit-ratio} presents the hit-ratio loading by the end of the procedure. 
For all items in all dimensions except one, we achieved a hit ratio above 50\%. We acknowledge a convergence drift in one of the items between comprehensibility and clarity. However, the wording of this item had a perfect alignment with the definition of comprehensibility, so we decided to keep it. Later analysis of the loading scores reinforced this decision. 
\chdeleted[id=del]{The resulting three items for each dimension grouped by construct are shown in}Figure~\ref{fig:explanation-quality-dimensions} and Figure~\ref{fig:background-factors}\chadded[id=add]{ denote the eventual wording for all scale items}.

\begin{table}[htb]
\Huge
\centering
\def\arraystretch{1.2}
\resizebox{\textwidth}{!}{%
\begin{tabular}{|l|lll|llll|llll|llll|}
\hline
\multirow{2}{*}{\textbf{\begin{tabular}[c]{@{}l@{}}Measurement\\ Dimension\end{tabular}}} & \multicolumn{3}{l|}{\textbf{\begin{tabular}[c]{@{}l@{}}Round 1\\ (5 judges)\end{tabular}}} & \multicolumn{4}{l|}{\textbf{\begin{tabular}[c]{@{}l@{}}Round 2 cumulative\\ (+5 judges)\end{tabular}}} & \multicolumn{4}{l|}{\textbf{\begin{tabular}[c]{@{}l@{}}Round 3 cumulative\\ (+5 judges)\end{tabular}}} & \multicolumn{4}{l|}{\textbf{\begin{tabular}[c]{@{}l@{}}Final\\ (15 judges)\end{tabular}}} \\ \cline{2-16} 
 & \multicolumn{1}{l|}{\textbf{\begin{tabular}[c]{@{}l@{}}\#\\ items\end{tabular}}} & \multicolumn{1}{l|}{\textbf{\begin{tabular}[c]{@{}l@{}}inter-\\ rater\\ agree\end{tabular}}} & \textbf{\begin{tabular}[c]{@{}l@{}}hit\\ ratio\end{tabular}} & \multicolumn{1}{l|}{\textbf{\begin{tabular}[c]{@{}l@{}}\#\\ items\end{tabular}}} & \multicolumn{1}{l|}{\textbf{\begin{tabular}[c]{@{}l@{}}\#\\ reworded\end{tabular}}} & \multicolumn{1}{l|}{\textbf{\begin{tabular}[c]{@{}l@{}}inter-\\ rater\\ agree\end{tabular}}} & \textbf{\begin{tabular}[c]{@{}l@{}}hit\\ ratio\end{tabular}} & \multicolumn{1}{l|}{\textbf{\begin{tabular}[c]{@{}l@{}}\#\\ items\end{tabular}}} & \multicolumn{1}{l|}{\textbf{\begin{tabular}[c]{@{}l@{}}\#\\ reworded\end{tabular}}} & \multicolumn{1}{l|}{\textbf{\begin{tabular}[c]{@{}l@{}}inter-\\ rater\\ agree\end{tabular}}} & \textbf{\begin{tabular}[c]{@{}l@{}}hit\\ ratio\end{tabular}} & \multicolumn{1}{l|}{\textbf{\begin{tabular}[c]{@{}l@{}}\# \\ items\end{tabular}}} & \multicolumn{1}{l|}{\textbf{\begin{tabular}[c]{@{}l@{}}\#\\ reworded\end{tabular}}} & \multicolumn{1}{l|}{\textbf{\begin{tabular}[c]{@{}l@{}}inter-\\ rater\\ agree\end{tabular}}} & \textbf{\begin{tabular}[c]{@{}l@{}}hit\\ ratio\end{tabular}} \\ \hline
Completeness & \multicolumn{1}{l|}{5} & \multicolumn{1}{l|}{0.58} & 0.72 & \multicolumn{1}{l|}{5} & \multicolumn{1}{l|}{2} & \multicolumn{1}{l|}{0.59} & 0.72 & \multicolumn{1}{l|}{5} & \multicolumn{1}{l|}{1} & \multicolumn{1}{l|}{0.45} & 0.61 & \multicolumn{1}{l|}{3} & \multicolumn{1}{l|}{1} & \multicolumn{1}{l|}{0.57} & 0.76 \\ \hline
Soundness & \multicolumn{1}{l|}{5} & \multicolumn{1}{l|}{0.46} & 0.6 & \multicolumn{1}{l|}{5} & \multicolumn{1}{l|}{3} & \multicolumn{1}{l|}{0.59} & 0.7 & \multicolumn{1}{l|}{5} & \multicolumn{1}{l|}{-} & \multicolumn{1}{l|}{0.42} & 0.59 & \multicolumn{1}{l|}{3} & \multicolumn{1}{l|}{-} & \multicolumn{1}{l|}{0.48} & 0.68 \\ \hline
Causability & \multicolumn{1}{l|}{5} & \multicolumn{1}{l|}{0.28} & 0.44 & \multicolumn{1}{l|}{5} & \multicolumn{1}{l|}{3} & \multicolumn{1}{l|}{0.38} & 0.6 & \multicolumn{1}{l|}{5} & \multicolumn{1}{l|}{1} & \multicolumn{1}{l|}{0.49} & 0.64 & \multicolumn{1}{l|}{3} & \multicolumn{1}{l|}{1} & \multicolumn{1}{l|}{0.68} & 0.81 \\ \hline
Clarity & \multicolumn{1}{l|}{5} & \multicolumn{1}{l|}{0.4} & 0.44 & \multicolumn{1}{l|}{5} & \multicolumn{1}{l|}{3} & \multicolumn{1}{l|}{0.43} & 0.64 & \multicolumn{1}{l|}{5} & \multicolumn{1}{l|}{1} & \multicolumn{1}{l|}{0.41} & 0.62 & \multicolumn{1}{l|}{3} & \multicolumn{1}{l|}{1} & \multicolumn{1}{l|}{0.49} & 0.69 \\ \hline
Compactness & \multicolumn{1}{l|}{5} & \multicolumn{1}{l|}{0.42} & 0.68 & \multicolumn{1}{l|}{5} & \multicolumn{1}{l|}{-} & \multicolumn{1}{l|}{0.4} & 0.6 & \multicolumn{1}{l|}{5} & \multicolumn{1}{l|}{2} & \multicolumn{1}{l|}{0.55} & 0.69 & \multicolumn{1}{l|}{3} & \multicolumn{1}{l|}{1} & \multicolumn{1}{l|}{0.70} & 0.82 \\ \hline
Comprehensibility & \multicolumn{1}{l|}{5} & \multicolumn{1}{l|}{0.16} & 0.32 & \multicolumn{1}{l|}{5} & \multicolumn{1}{l|}{4} & \multicolumn{1}{l|}{0.4} & 0.38 & \multicolumn{1}{l|}{5} & \multicolumn{1}{l|}{2} & \multicolumn{1}{l|}{0.38} & 0.40 & \multicolumn{1}{l|}{3} & \multicolumn{1}{l|}{-} & \multicolumn{1}{l|}{0.30} & 0.47 \\ \hline
Curiosity & \multicolumn{1}{l|}{5} & \multicolumn{1}{l|}{0.78} & 0.88 & \multicolumn{1}{l|}{5} & \multicolumn{1}{l|}{-} & \multicolumn{1}{l|}{0.76} & 0.86 & \multicolumn{1}{l|}{5} & \multicolumn{1}{l|}{-} & \multicolumn{1}{l|}{0.79} & 0.88 & \multicolumn{1}{l|}{3} & \multicolumn{1}{l|}{-} & \multicolumn{1}{l|}{0.91} & 0.96 \\ \hline
Trust & \multicolumn{1}{l|}{5} & \multicolumn{1}{l|}{0.72} & 0.84 & \multicolumn{1}{l|}{5} & \multicolumn{1}{l|}{-} & \multicolumn{1}{l|}{0.76} & 0.86 & \multicolumn{1}{l|}{5} & \multicolumn{1}{l|}{-} & \multicolumn{1}{l|}{0.76} & 0.85 & \multicolumn{1}{l|}{3} & \multicolumn{1}{l|}{-} & \multicolumn{1}{l|}{0.91} & 0.96 \\ \hline
\end{tabular}%
}
\caption{Card sorting results: cumulative inter-rater agreement and hit ratio }
\label{tab:card-sorting-result}
\end{table}

\begin{table}[htb]
\large
\centering
\def\arraystretch{1.1}
\resizebox{\textwidth}{!}{%
\begin{tabular}{|
>{\columncolor[HTML]{FFFFFF}}l |l|l|l|l|l|l|l|l|}
\hline
{\color[HTML]{000000} } & \cellcolor[HTML]{FFFFFF}{\color[HTML]{333333} \textbf{Causality}} & \cellcolor[HTML]{FFFFFF}{\color[HTML]{333333} \textbf{Clarity}} & \cellcolor[HTML]{FFFFFF}{\color[HTML]{333333} \textbf{Compactness}} & \cellcolor[HTML]{FFFFFF}{\color[HTML]{333333} \textbf{Completeness}} & \cellcolor[HTML]{FFFFFF}{\color[HTML]{333333} \textbf{Comprehensibility}} & \cellcolor[HTML]{FFFFFF}{\color[HTML]{333333} \textbf{Curiosity}} & \cellcolor[HTML]{FFFFFF}{\color[HTML]{333333} \textbf{Soundness}} & \cellcolor[HTML]{FFFFFF}{\color[HTML]{333333} \textbf{Trust}} \\ \hline
{\color[HTML]{000000} COM-Q1} & 20\% &  &  & \cellcolor[HTML]{404040}{\color[HTML]{FFFFFF} 80\%} &  &  &  &  \\ \hline
{\color[HTML]{000000} COM-Q2} &  &  &  & \cellcolor[HTML]{404040}{\color[HTML]{FFFFFF} 80\%} & 13\% &  &  & 7\% \\ \hline
{\color[HTML]{000000} COM-Q3} &  &  & 7\% & \cellcolor[HTML]{595959}{\color[HTML]{FFFFFF} 67\%} & 20\% &  &  & 7\% \\ \hline
{\color[HTML]{000000} SND-Q1} &  & 7\% &  &  & 7\% &  & \cellcolor[HTML]{404040}{\color[HTML]{FFFFFF} 73\%} & 13\% \\ \hline
{\color[HTML]{000000} SND-Q2} &  & 7\% &  &  &  &  & \cellcolor[HTML]{808080}{\color[HTML]{FFFFFF} 60\%} & 33\% \\ \hline
{\color[HTML]{000000} SND-Q3} &  & 10\% &  & 10\% & 10\% &  & \cellcolor[HTML]{595959}{\color[HTML]{FFFFFF} 70\%} &  \\ \hline
{\color[HTML]{000000} CAS-Q1} & \cellcolor[HTML]{595959}{\color[HTML]{FFFFFF} 70\%} &  &  & 10\% &  &  & 20\% &  \\ \hline
{\color[HTML]{000000} CAS-Q2} & \cellcolor[HTML]{0D0D0D}{\color[HTML]{FFFFFF} 100\%} &  &  &  &  &  &  &  \\ \hline
{\color[HTML]{000000} CAS-Q3} & \cellcolor[HTML]{404040}{\color[HTML]{FFFFFF} 73\%} & 20\% &  &  &  &  & 7\% &  \\ \hline
{\color[HTML]{000000} CLR-Q1} &  & \cellcolor[HTML]{595959}{\color[HTML]{FFFFFF} 67\%} & 7\% &  & 20\% &  & 7\% &  \\ \hline
{\color[HTML]{000000} CLR-Q2} & 20\% & \cellcolor[HTML]{808080}{\color[HTML]{FFFFFF} 60\%} &  &  &  &  & 20\% &  \\ \hline
{\color[HTML]{000000} CLR-Q3} &  & \cellcolor[HTML]{404040}{\color[HTML]{FFFFFF} 80\%} & 10\% &  & 10\% &  &  &  \\ \hline
{\color[HTML]{000000} CMP-Q1} &  &  & \cellcolor[HTML]{0D0D0D}{\color[HTML]{FFFFFF} 100\%} &  &  &  &  &  \\ \hline
{\color[HTML]{000000} CMP-Q2} &  & 7\% & \cellcolor[HTML]{595959}{\color[HTML]{FFFFFF} 67\%} & 7\% &  &  & 20\% &  \\ \hline
{\color[HTML]{000000} CMP-Q3} &  &  & \cellcolor[HTML]{404040}{\color[HTML]{FFFFFF} 80\%} & 13\% &  &  & 7\% &  \\ \hline
{\color[HTML]{000000} CPR-Q1} & 10\% & 10\% &  &  & \cellcolor[HTML]{A6A6A6}50\% &  & 10\% & 20\% \\ \hline
{\color[HTML]{000000} CPR-Q2} &  & \cellcolor[HTML]{BFBFBF}40\% & 30\% &  & \cellcolor[HTML]{D9D9D9}30\% &  &  &  \\ \hline
{\color[HTML]{000000} CPR-Q3} &  & 20\% &  &  & \cellcolor[HTML]{808080}{\color[HTML]{FFFFFF} 60\%} &  & 13\% & 7\% \\ \hline
{\color[HTML]{000000} CUR-Q1} &  &  &  &  &  & \cellcolor[HTML]{0D0D0D}{\color[HTML]{FFFFFF} 100\%} &  &  \\ \hline
{\color[HTML]{000000} CUR-Q2} &  &  &  &  &  & \cellcolor[HTML]{0D0D0D}{\color[HTML]{FFFFFF} 100\%} &  &  \\ \hline
{\color[HTML]{000000} CUR-Q3} &  & 7\% &  & 7\% &  & \cellcolor[HTML]{262626}{\color[HTML]{FFFFFF} 87\%} &  &  \\ \hline
{\color[HTML]{000000} TRU-Q1} &  &  &  &  &  &  &  & \cellcolor[HTML]{0D0D0D}{\color[HTML]{FFFFFF} 100\%} \\ \hline
{\color[HTML]{000000} TRU-Q2} &  & 7\% &  & 7\% &  &  &  & \cellcolor[HTML]{262626}{\color[HTML]{FFFFFF} 87\%} \\ \hline
{\color[HTML]{000000} TRU-Q3} &  &  &  &  &  &  &  & \cellcolor[HTML]{0D0D0D}{\color[HTML]{FFFFFF} 100\%} \\ \hline
\end{tabular}%
}
\caption{Item hit-ratio after the sorting procedure}
\label{tab:card-sorting-hit-ratio}
\end{table}


\subsection {User Study}

In our experiment, we primarily focused on investigating how different types of explanatory knowledge about BPs could be incorporated into the LLM prompt preceding the interaction with the LLMs when seeking explanations about process execution outcomes. As an independent variable in our experimental design, we employed three between-group manipulations based on distinct knowledge types: process, XAI (feature importance), and causal. All input combinations are given in Table~\ref{tab:knowledge-manipulations}.

We consistently used BP knowledge as a baseline in all manipulations. The first manipulation involved augmenting this baseline with XAI knowledge related to the factors influencing the specific process condition being explained. This was compared to using only XAI knowledge as input. 
\chdeleted[id=del]{In the second manipulation, instead of XAI knowledge, we explored the impact of adding causal knowledge about the execution dependencies among process activities that could lead to the condition being explained. This was compared with using only the baseline process knowledge as an input. Lastly, the third manipulation compared the effect of combining process and XAI knowledge with the effect of integrating all three knowledge types as input for the LLM.}
\chadded[id=add]{In the second manipulation, we compared the effect of combining process and XAI knowledge with the effect of integrating all three knowledge types as input for the LLM.
Lastly, instead of XAI knowledge, we explored the impact of adding causal knowledge about the execution dependencies among process activities that could lead to the condition being explained. This was compared with using only the baseline process knowledge as an input.}

\begin{table}[htb]
    \centering
    \def\arraystretch{1.2}%
    \Large
    \resizebox{\textwidth}{!}{\begin{tabular}{*{3}{l}}
    \hline
    \textbf{Problem Domain} & \textbf{Group 1} & \textbf{Group 2} \\
    \hline
    Pizza delivery & Process and Feature-Importance (XAI) & Feature-Importance (XAI) \\
    \hline
    Parking fines & Process and Feature-Importance (XAI) & Process, Feature-Importance (XAI), and Causal\\
    \hline
    Loan approval & Process and Causal & Process \\
    \hline
    \end{tabular}}
    \caption{Three manipulations of input knowledge type}
    \label{tab:knowledge-manipulations}
\end{table}

Respective to the various manipulations depicted in Table~\ref{tab:knowledge-manipulations}, each of the above hypotheses was further instantiated for more concrete testing as follows:

\newtheorem{subhyp}{Subhypothesis}

\FloatBarrier
\begin{subhyp}[H\ref{hyp:first}.1]
\label{subhyp:first}
Explanations generated by LLMs informed by knowledge about \ul{business processes and feature importance} (XAI) will be perceived as having higher \textbf{fidelity} compared to explanations informed \ul{only by feature importance knowledge} (XAI).
\end{subhyp}

\begin{subhyp}[H\ref{hyp:first}.2]
\label{subhyp:second}
Explanations generated by LLMs informed by knowledge about \ul{business processes, feature importance, and causal execution dependencies} will be perceived as having higher \textbf{fidelity} compared to explanations informed \ul{only by business processes and feature importance knowledge}.
\end{subhyp}

\begin{subhyp}[H\ref{hyp:first}.3]
\label{subhyp:third}
Explanations generated by LLMs informed by knowledge about \ul{business processes and causal execution dependencies} will be perceived as having higher \textbf{fidelity} compared to explanations informed \ul{only by BP knowledge}.
\end{subhyp}

\begin{subhyp}[H\ref{hyp:second}.1]
\label{subhyp:fourth}
Explanations generated by LLMs informed by knowledge about \ul{business processes and feature importance} (XAI) will be perceived as having higher \textbf{interpretability} compared to explanations informed \ul{only by feature importance knowledge} (XAI).
\end{subhyp}

\begin{subhyp}[H\ref{hyp:second}.2]
\label{subhyp:fifth}
Explanations generated by LLMs informed by knowledge about \ul{business processes, feature importance, and causal execution dependencies} will be perceived as having higher \textbf{interpretability} compared to explanations informed \ul{only by business processes and feature importance knowledge}.
\end{subhyp}

\begin{subhyp}[H\ref{hyp:second}.3]
\label{subhyp:sixth}
Explanations generated by LLMs informed by knowledge about \ul{business processes and causal execution dependencies} will be perceived as having higher \textbf{interpretability} compared to explanations informed \ul{only by BP knowledge}.
\end{subhyp}
\FloatBarrier
\subsection {Domains}

Each manipulation was implemented within a distinct problem domain to derive explanations for specific conditions. In particular, the first manipulation, hypothesized in H1.1 and H2.1, was developed within the `pizza delivery' domain, focusing on creating explanations for delays in pizza delivery. The second manipulation, outlined in H1.2 and H2.2, is related to the `parking fines' domain, targeting the development of explanations for the lateness in fine processing. Finally, the third manipulation, as hypothesized in H1.3 and H2.3, was situated in the `loan approval' domain, with a focus on elucidating the potential reduction in loan approval times.



Corresponding to each domain, we generated the knowledge ingredients about the process model, the causal model, and the XAI feature importance using the SAX4BPM library services. This was applied to two synthesized process data in the pizza delivery and parking fines domains, and to the loan approval domain using an open dataset\footnote{\url{https://data.4tu.nl/articles/dataset/BPI_Challenge_2017/12696884}}.




\section{Experimental Design}
\label{sec:experimental-design}

\begin{figure}[!ht]
    \centering
    \includegraphics[width=1\linewidth]{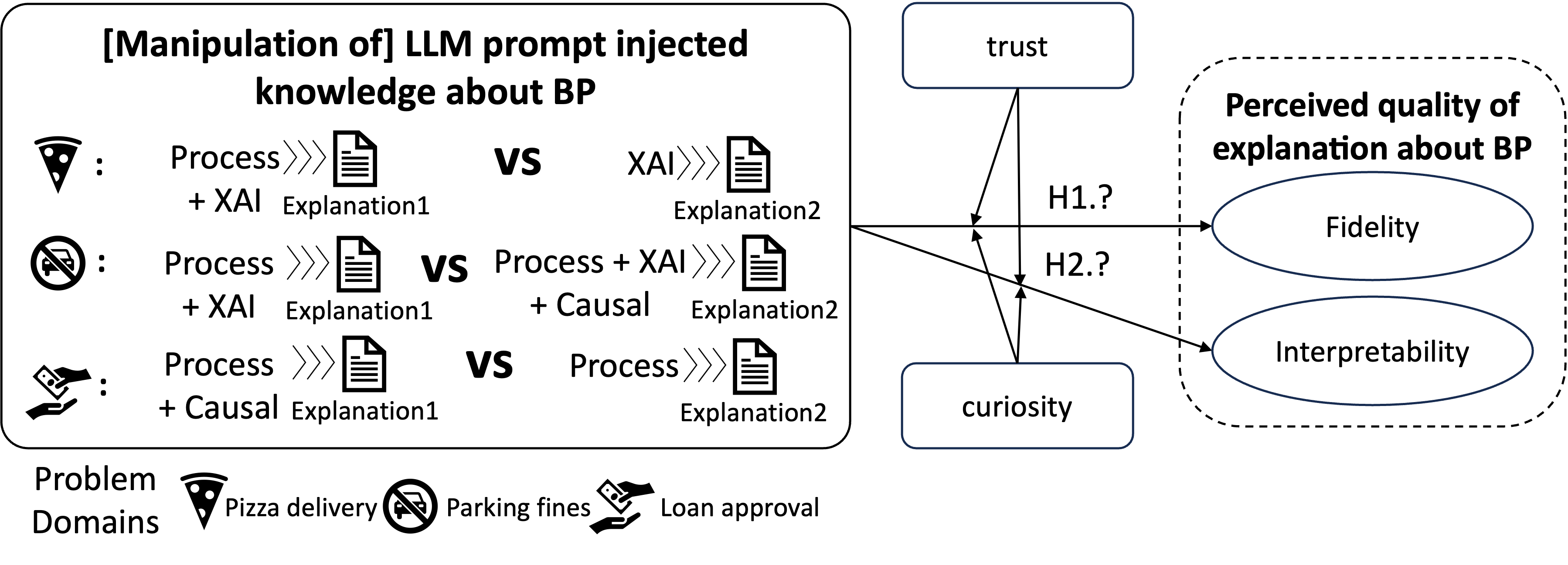}
    \caption{Experiment model}
    \label{fig:experiment-model}
\end{figure}

We conducted a 3X2 (knowledge type, perceived explanation quality) mixed experimental design to evaluate our hypotheses. This design was applied across the three problem domains\chdeleted[id=del]{: pizza delivery, parking fines, and loan approval}. In the experimental setup, participants were randomly assigned to one of two distinct groups. They completed an online questionnaire that included a consent form, basic demographics, and a series of 3 sets of [1-7 Likert] rating scales relevant to the aforementioned problem domains. The Likert method~\cite{Likert1932} is widely used as a standard psychometric scale to measure human responses. Each set comprised 24 items (see Figures~\ref{fig:explanation-quality-dimensions} and \ref{fig:background-factors}), phrased based on the methodological scale development process described in section~\ref{sec:scale-development}. This included 9 items each for fidelity and interpretability (3 for each dimension), plus 6 additional items (3 each for curiosity and trust). 
Preceding each set of rating scales was a description of a specific BP in one domain, a question addressing a particular process condition requiring an explanation, a “ground truth” explanation provided by the researchers, and an explanation generated by the LLM (in our case \chadded[id=add]{GPT4.0}\chdeleted[id=del]{ChatGPT 4}) in response to the same question. 
The ground truth served as a baseline to which we can make the comparisons relative. 
Participants were instructed to rate the LLM-generated explanation, assessing its quality against the ground truth. Each LLM \chdeleted[id=del]{was pre-prompted}\chadded[id=add]{prompt consisted of a} descriptive text\chadded[id=add]{, which was a blend of} \chdeleted[id=del]{blending} the manipulated knowledge elements, as detailed in Figure~\ref{fig:experiment-model}. The two questionnaires can be accessed at:~\url{https://github.com/IBM/SAX/tree/main/KDE-SI-2024/survey}.

In the pizza-delivery case, the explanation concerned possible reasons for pizza delivery lateness, and the explanation generation by the LLM varied between prompt inputs that included both process and XAI knowledge (group 1) and prompt inputs that included only XAI knowledge (group 2). In the parking-fines case, the explanation addressed possible reasons for the extensive processing time of parking fines. Here, the LLM-generated explanation differed between prompt inputs that included process, causal, and XAI knowledge (group 1) and prompt inputs that included process and XAI knowledge (group 2). Lastly, in the loan-approval case, the explanation focused on reasons for delayed loan offers. Accordingly, the explanation generated by the LLM was modified between prompt inputs that included both process and causal knowledge (group 1) and prompt inputs that included only process knowledge (group 2).

\chadded[id=add]{We emphasize that with regards to correctness, our experimental protocol presented the users also with the ground truth version of the explanations to eliminate reliance on incorrect output. Moreover, the analysis of the effect of the different knowledge ingredients is comparative rather than absolute. To a certain extent, the result of the empirical evaluation pursued in this work is also evidence of the correctness of the output from a perceived standpoint.}

\section{Results}
\label{sec:results}

\subsection{Data Preparation} 

Before running the statistical analysis, we explored the data using BOXPLOTs for the two target constructs across all three domains in order to determine extreme responses. There were only 2 responses out of the 49 (1 for fidelity in the parking-fines domain, and 1 for \chdeleted[id=del]{interoperability}\chadded[id=add]{interpretability} in the loan-approval) that were mild outliers, scoring more than 1.5 times the inter-quartile range\footnote{The inter-quartile range (IQR) contains the middle 50 percent of the distribution. For normally distributed data, a range that is 1.5 times the IQR covers $\sim99.3\%$ of the distribution population.}. These two responses were not consistent across the three domains, hence we did not remove them from the analysis. Running a similar BOXPLOT analysis for general homogeneity of variance across all individual respondents yielded two respondents who were identified as having mildly low variance in their responses overall. However, we executed our subsequent analysis for the main effects twice, with and without the outliers, concluding that outlier elimination is negligible and has no effect on the significance of our results. Hence, the final results reporting utilized 47 responses across all three domains.

\begin{table}[ht]
\centering
\def\arraystretch{1.2}%
\resizebox{\textwidth}{!}{%
\begin{tabular}{|l|l|l|l|}
\hline
\textbf{Measurement dimension} & \textbf{Cronbach's Alpha} & \textbf{Construct} & \textbf{Cronbach's alpha} \\ \hline
Completeness & 0.91 &  &  \\ \cline{1-2}
Soundness & 0.63 &  &  \\ \cline{1-2}
Causability & 0.69 & \multirow{-3}{*}{Fidelity} & \multirow{-3}{*}{0.81} \\ \hline
Clarity & 0.81 &  &  \\ \cline{1-2}
Compactness & 0.49 (0.516) &  &  \\ \cline{1-2}
Comprehensibility & 0.64 & \multirow{-3}{*}{Interpretebility} & \multirow{-3}{*}{0.80} \\ \hline
Curiosity & 0.92 & \cellcolor[HTML]{9B9B9B} & \cellcolor[HTML]{9B9B9B} \\ \hline
Trust & 0.90 & \cellcolor[HTML]{9B9B9B} & \cellcolor[HTML]{9B9B9B} \\ \hline
\end{tabular}%
}
\caption{Reliability scores}
\label{tab:reliability}
\end{table} 

\subsection{Scale Reliability} 

After exploring outliers, we assessed the scale reliability for all measurement dimensions. The reliability scores, item-wise (three per dimension), for each measurement dimension and the latent constructs, are presented in Table~\ref{tab:reliability}. Cronbach's alpha values ranging from 0.6 to 0.8 are conventionally acceptable~\cite{Shi2012ContentDevelopment.}. As shown in the table, most measurement dimensions had Cronbach's alpha scores above 0.6 (ranging from 0.63 to 0.92), indicating an overall internal consistency of 0.81 for fidelity and 0.80 for \chdeleted[id=del]{interoperability}\chadded[id=add]{interpretability}. This excludes the `compactness' dimension, which scored below the recommended threshold with an alpha of 0.49. Eliminating the third item from this dimension slightly improved its internal consistency to 0.516. To further validate these findings, we conducted an exploratory factor analysis for all measurement dimensions, revealing a single component for all dimensions with factor loading scores as shown in Table~\ref{tab:factor-loadings}. The results were consistent with our reliability scores, with most loading scores surpassing the recommended loading threshold (> 0.4, \cite{Streiner1994FiguringAnalysis.}) and explained variance exceeding 50\%. Notably, the loadings for the compactness dimension, particularly its third item, were marginally above the threshold (0.584). Therefore, we repeated the analyses for the main effects both with and without this item. Its removal did not significantly alter the results, eventually leading to our decision to retain it.

\begin{table}[ht]
\centering
\footnotesize
\def\arraystretch{1}%
\resizebox{\textwidth}{!}{%
\begin{tabular}{|l|l|l|l|l|}
\hline
\textbf{Measurement dimension} & \textbf{Item 1} & \textbf{Item 2} & \textbf{Item 3} & \textbf{Variance explained} \\ \hline
Completeness & 0.919 & 0.924 & 0.914 & 84.42\% \\ \hline
Soundness & 0.837 & 0.606 & 0.821 & 58.05\% \\ \hline
Causability & 0.829 & 0.804 & 0.718 & 61.61\% \\ \hline
Clarity & 0.846 & 0.863 & 0.841 & 72.24\% \\ \hline
Compactness & 0.714 & 0.804 & 0.584 & 49.88\% \\ \hline
Comprehensibility & 0.803 & 0.667 & 0.835 & 59.53\% \\ \hline
Curiosity & 0.930 & \cellcolor[HTML]{FFFFFF}0.955 & \cellcolor[HTML]{FFFFFF}0.894 & 85.88\% \\ \hline
Trust & 0.934 & \cellcolor[HTML]{FFFFFF}0.869 & \cellcolor[HTML]{FFFFFF}0.928 & 82.94\% \\ \hline
\end{tabular}%
}
\caption{Factor loadings}
\label{tab:factor-loadings}
\end{table}

\begin{table}[ht]
\centering
\def\arraystretch{1.2}%
\resizebox{\textwidth}{!}{%
\begin{tabular}{|l|l|l|lll|lll|}
\hline
\multirow{2}{*}{\textbf{Problem domain}} & \multirow{2}{*}{\textbf{Construct}} & \multirow{2}{*}{\textbf{Measurement dimension}} & \multicolumn{3}{c|}{\textbf{Group 1}} & \multicolumn{3}{c|}{\textbf{Group 2}} \\ \cline{4-9} 
 &  &  & \multicolumn{1}{l|}{\textbf{M}} & \multicolumn{1}{l|}{\textbf{SD}} & \textbf{n} & \multicolumn{1}{l|}{\textbf{M}} & \multicolumn{1}{l|}{\textbf{SD}} & \textbf{n} \\ \hline
\multirow{6}{*}{Pizza delivery} & \multirow{3}{*}{Fidelity} & Completeness & \multicolumn{1}{l|}{4.48} & \multicolumn{1}{l|}{1.58} & 24 & \multicolumn{1}{l|}{3.73} & \multicolumn{1}{l|}{1.92} & 23 \\ \cline{3-9} 
 &  & Soundness & \multicolumn{1}{l|}{4.31} & \multicolumn{1}{l|}{0.98} & 24 & \multicolumn{1}{l|}{3.59} & \multicolumn{1}{l|}{1.13} & 23 \\ \cline{3-9} 
 &  & Causability & \multicolumn{1}{l|}{4.16} & \multicolumn{1}{l|}{1.39} & 24 & \multicolumn{1}{l|}{3.65} & \multicolumn{1}{l|}{1.08} & 23 \\ \cline{2-9} 
 & \multirow{3}{*}{Interpretability} & Clarity & \multicolumn{1}{l|}{4.37} & \multicolumn{1}{l|}{1.20} & 24 & \multicolumn{1}{l|}{4.69} & \multicolumn{1}{l|}{1.42} & 23 \\ \cline{3-9} 
 &  & Compactness & \multicolumn{1}{l|}{3.50} & \multicolumn{1}{l|}{1.13} & 24 & \multicolumn{1}{l|}{4.10} & \multicolumn{1}{l|}{1.19} & 23 \\ \cline{3-9} 
 &  & Comprehensibility & \multicolumn{1}{l|}{4.08} & \multicolumn{1}{l|}{1.29} & 24 & \multicolumn{1}{l|}{3.71} & \multicolumn{1}{l|}{1.26} & 23 \\ \hline
\multirow{6}{*}{Parking fines} & \multirow{3}{*}{Fidelity} & Completeness & \multicolumn{1}{l|}{2.79} & \multicolumn{1}{l|}{1.56} & 24 & \multicolumn{1}{l|}{3.65} & \multicolumn{1}{l|}{1.66} & 23 \\ \cline{3-9} 
 &  & Soundness & \multicolumn{1}{l|}{3.18} & \multicolumn{1}{l|}{0.96} & 24 & \multicolumn{1}{l|}{3.62} & \multicolumn{1}{l|}{1.10} & 23 \\ \cline{3-9} 
 &  & Causability & \multicolumn{1}{l|}{3.72} & \multicolumn{1}{l|}{1.38} & 24 & \multicolumn{1}{l|}{3.88} & \multicolumn{1}{l|}{1.12} & 23 \\ \cline{2-9} 
 & \multirow{3}{*}{Interpretability} & Clarity & \multicolumn{1}{l|}{4.25} & \multicolumn{1}{l|}{1.68} & 24 & \multicolumn{1}{l|}{2.73} & \multicolumn{1}{l|}{1.32} & 23 \\ \cline{3-9} 
 &  & Compactness & \multicolumn{1}{l|}{3.23} & \multicolumn{1}{l|}{1.12} & 24 & \multicolumn{1}{l|}{2.42} & \multicolumn{1}{l|}{0.95} & 23 \\ \cline{3-9} 
 &  & Comprehensibility & \multicolumn{1}{l|}{3.41} & \multicolumn{1}{l|}{1.20} & 24 & \multicolumn{1}{l|}{2.53} & \multicolumn{1}{l|}{0.91} & 23 \\ \hline
\multirow{6}{*}{Loan approval} & \multirow{3}{*}{Fidelity} & Completeness & \multicolumn{1}{l|}{3.66} & \multicolumn{1}{l|}{1.54} & 24 & \multicolumn{1}{l|}{3.63} & \multicolumn{1}{l|}{1.18} & 23 \\ \cline{3-9} 
 &  & Soundness & \multicolumn{1}{l|}{4.31} & \multicolumn{1}{l|}{0.98} & 24 & \multicolumn{1}{l|}{3.59} & \multicolumn{1}{l|}{1.13} & 23 \\ \cline{3-9} 
 &  & Causability & \multicolumn{1}{l|}{4.16} & \multicolumn{1}{l|}{1.39} & 24 & \multicolumn{1}{l|}{3.65} & \multicolumn{1}{l|}{1.08} & 23 \\ \cline{2-9} 
 & \multirow{3}{*}{Interpretability} & Clarity & \multicolumn{1}{l|}{3.31} & \multicolumn{1}{l|}{1.62} & 24 & \multicolumn{1}{l|}{3.11} & \multicolumn{1}{l|}{1.26} & 23 \\ \cline{3-9} 
 &  & Compactness & \multicolumn{1}{l|}{3.18} & \multicolumn{1}{l|}{1.02} & 24 & \multicolumn{1}{l|}{3.31} & \multicolumn{1}{l|}{1.02} & 23 \\ \cline{3-9} 
 &  & Comprehensibility & \multicolumn{1}{l|}{3.48} & \multicolumn{1}{l|}{1.28} & 24 & \multicolumn{1}{l|}{3.31} & \multicolumn{1}{l|}{1.14} & 23 \\ \hline
\end{tabular}%
}
\caption{Descriptives for all measurement dimensions across groups }
\label{tab:dimension-descriptives}
\end{table}

\subsection{Effect on perceived explanation quality}

Descriptive statistics for all underlying measurement dimensions are detailed in Table~\ref{tab:dimension-descriptives}. Domain-wise analysis of the effects was carried out. 

\subsubsection{Pizza-delivery domain}
Results for the pizza-delivery domain are detailed in Table~\ref{tab:main-effect-pizza}.
An Analysis of Variance (ANOVA) was conducted first to examine the effects of knowledge manipulation (between-group differences) on two dependent variables: fidelity and interpretability. The analysis revealed that for the dependent variable of fidelity, there was no significant effect, F(1,46) = 2.42, p = .12, $\eta^2$ = .051. This suggests that knowledge type alone does not have a substantial effect on fidelity. Also for the dependent variable of interpretability, the effect was not statistically significant, F(1, 46) = 0.42, p = .51, $\eta^2$ = .009, suggesting that the type of knowledge manipulation that was conducted has no significant effect on interpretability when not controlling for any covariate.

An Analysis of Covariance (ANCOVA) was conducted next to evaluate the effects of knowledge type manipulation on the two dependent variables: fidelity and interpretability while controlling for the covariates trust and curiosity. The analysis revealed that for the dependent variable fidelity, the adjusted model was statistically significant, F(3, 44) = 7.60, p < .001, $\eta^2$ = .347, indicating a substantial effect of the controlled manipulation. Similarly, regarding the dependent variable interpretability, the adjusted model was also significant, F(3, 44) = 5.08, p = .004, $\eta^2$ = .262. Thus, when incorporating trust and curiosity as covariates, the manipulation of knowledge type as was realized in the pizza-delivery domain, significantly affected the perceptions of fidelity and \chdeleted[id=del]{interoperability}\chadded[id=add]{interpretability}.

Running the ANCOVA analysis also revealed that trust has a significant direct effect on both dependent variables: on fidelity, F(1,46) = 17.79, p < .001, $\eta^2$ = .29, and on interpretability, F(1,46) = 14.67, p < .001, $\eta^2$ = .25.

Therefore, in relation to the corresponding hypotheses (H1.1; H2.1), the results support the expectation that adding process-related knowledge to XAI knowledge as an input to the LLM can potentially yield a significant effect on both perceptions of fidelity and \chdeleted[id=del]{interoperability}\chadded[id=add]{interpretability}. However, this effect is moderated by the degree to which the user is curious about the condition explored and trusts the LLM tool. Stretched beyond our original hypotheses, the results also identified trust as a background factor. This factor not only moderates the effect of the manipulated input to the LLM but could also directly mask this effect.

\begin{table}[ht]
\centering
\def\arraystretch{1.2}%
\Large
\resizebox{\textwidth}{!}{%
\begin{tabular}{|l|llllllll|llllllll|}
\hline
\multirow{3}{*}{\textbf{Construct}} & \multicolumn{8}{l|}{\textbf{ANOVA}} & \multicolumn{8}{l|}{\textbf{ANCOVA}} \\ \cline{2-17} 
 & \multicolumn{2}{l|}{\textbf{Group 1}} & \multicolumn{2}{l|}{\textbf{Group 2}} & \multicolumn{1}{l|}{\multirow{2}{*}{\textbf{df}}} & \multicolumn{1}{l|}{\multirow{2}{*}{\textbf{F}}} & \multicolumn{1}{l|}{\multirow{2}{*}{\textbf{Sig.}}} & \multirow{2}{*}{\textbf{$\eta^2$}} & \multicolumn{2}{l|}{\textbf{Group 1}} & \multicolumn{2}{l|}{\textbf{Group 2}} & \multicolumn{1}{l|}{\multirow{2}{*}{\textbf{df}}} & \multicolumn{1}{l|}{\multirow{2}{*}{\textbf{F}}} & \multicolumn{1}{l|}{\multirow{2}{*}{\textbf{Sig.}}} & \multirow{2}{*}{\textbf{$\eta^2$}} \\ \cline{2-5} \cline{10-13}
 & \multicolumn{1}{l|}{M} & \multicolumn{1}{l|}{SE} & \multicolumn{1}{l|}{M} & \multicolumn{1}{l|}{SE} & \multicolumn{1}{l|}{} & \multicolumn{1}{l|}{} & \multicolumn{1}{l|}{} &  & \multicolumn{1}{l|}{M} & \multicolumn{1}{l|}{SE} & \multicolumn{1}{l|}{M} & \multicolumn{1}{l|}{SE} & \multicolumn{1}{l|}{} & \multicolumn{1}{l|}{} & \multicolumn{1}{l|}{} &  \\ \hline
Fidelity & \multicolumn{1}{l|}{4.44} & \multicolumn{1}{l|}{0.26} & \multicolumn{1}{l|}{3.86} & \multicolumn{1}{l|}{0.26} & \multicolumn{1}{l|}{(1,46)} & \multicolumn{1}{l|}{2.42} & \multicolumn{1}{l|}{0.12} & 0.051 & \multicolumn{1}{l|}{4.46} & \multicolumn{1}{l|}{0.22} & \multicolumn{1}{l|}{3.84} & \multicolumn{1}{l|}{0.22} & \multicolumn{1}{l|}{(3,44)} & \multicolumn{1}{l|}{7.60} & \multicolumn{1}{l|}{\textless{}.001*} & 0.347 \\ \hline
Interpretability & \multicolumn{1}{l|}{3.98} & \multicolumn{1}{l|}{0.19} & \multicolumn{1}{l|}{4.16} & \multicolumn{1}{l|}{0.20} & \multicolumn{1}{l|}{(1,46)} & \multicolumn{1}{l|}{0.42} & \multicolumn{1}{l|}{0.51} & 0.009 & \multicolumn{1}{l|}{4.01} & \multicolumn{1}{l|}{0.17} & \multicolumn{1}{l|}{4.13} & \multicolumn{1}{l|}{0.17} & \multicolumn{1}{l|}{(3,44)} & \multicolumn{1}{l|}{5.08} & \multicolumn{1}{l|}{.004*} & 0.262 \\ \hline
\end{tabular}%
}
\caption{ANOVA and ANCOVA (i.e., including curiosity and trust as covariates) results for the pizza delivery case. When controlling for curiosity and trust,
both effects on fidelity and \chdeleted[id=del]{interoperability}\chadded[id=add]{interpretability} were significant.}
\label{tab:main-effect-pizza}
\end{table}

\subsubsection{Parking fines domain}

Results for the parking-fines domain are detailed in Table~\ref{tab:main-effect-parking}.
For the parking-fine case, an \chdeleted[id=del]{Analysis of Variance (}ANOVA\chdeleted[id=del]{)} \chadded[id=add]{analysis} to examine the effects of knowledge manipulation on two dependent variables fidelity and \chdeleted[id=del]{interoperability}\chadded[id=add]{interpretability}, revealed that for the dependent variable of fidelity, there was no significant effect, F(1,46) = 2.71, p = .107, $\eta^2$ = .057. However, for the dependent variable of interpretability, the effect was statistically significant, F(1, 46) = 15.30, p = <.001, $\eta^2$ = .254, suggesting that the particular type of knowledge manipulation that was conducted in this case has a significant effect on interpretability, even when not controlling for any covariate.

\chdeleted[id=del]{An Analysis of Covariance (ANCOVA) was conducted next to evaluate the effects of knowledge manipulation on the two dependent variables while controlling for the covariates trust and curiosity. The }\chadded[id=add]{ANCOVA} analysis revealed that for the dependent variable fidelity, the adjusted model was statistically significant, F(3, 44) = 3.67, p = .019,
$\eta^2$ = .204, indicating a substantial effect of the manipulation. Similarly, regarding the dependent variable interpretability, the adjusted model was also significant, F(3, 44) = 9.84, p = <.001, $\eta^2$ = .407. Thus, when incorporating trust and curiosity as covariates, the manipulation of knowledge type as was realized in the parking-fines domain, significantly affected the perceptions of fidelity and \chdeleted[id=del]{interoperability}\chadded[id=add]{interpretability} while controlling for trust and curiosity. For interpretability, the effect size is 15\% stronger than when not controlling for the covariates.

Running the ANCOVA analysis also revealed that curiosity has a significant direct effect on the perception of interpretability, F(1,46) = 6.43, p = 0.015, $\eta^2$ = 0.13, and as in the pizza domain, trust has a significant effect on both fidelity, F(1,46) = 5.71, p = .021, $\eta^2$ = .11, and on interpretability, F(1,46) = 7.00, p = .011, $\eta^2$ = .14.

Concerning the corresponding hypotheses (H1.2; H2.2), the results support the expectation that adding causal-related knowledge to process and XAI knowledge as an input to the LLM can potentially yield a significant effect on both perceptions of fidelity and \chdeleted[id=del]{interoperability}\chadded[id=add]{interpretability}. In the case of interpretability, the effect is present regardless of any particular moderation, whereas in the case of fidelity, the effect is moderated by the user's curiosity about the explored condition and trust in the LLM tool. Beyond our original hypotheses, it is also observed that curiosity has a direct effect on interpretability, and trust has a direct effect on both perceptions, interpretability, and fidelity. 

\begin{table}[ht]
\centering
\def\arraystretch{1.2}%
\Large
\resizebox{\textwidth}{!}{%
\begin{tabular}{|l|llllllll|llllllll|}
\hline
\multirow{3}{*}{\textbf{Construct}} & \multicolumn{8}{l|}{\textbf{ANOVA}} & \multicolumn{8}{l|}{\textbf{ANCOVA}} \\ \cline{2-17} 
 & \multicolumn{2}{l|}{\textbf{Group 1}} & \multicolumn{2}{l|}{\textbf{Group 2}} & \multicolumn{1}{l|}{\multirow{2}{*}{\textbf{df}}} & \multicolumn{1}{l|}{\multirow{2}{*}{\textbf{F}}} & \multicolumn{1}{l|}{\multirow{2}{*}{\textbf{Sig.}}} & \multirow{2}{*}{\textbf{$\eta^2$}} & \multicolumn{2}{l|}{\textbf{Group 1}} & \multicolumn{2}{l|}{\textbf{Group 2}} & \multicolumn{1}{l|}{\multirow{2}{*}{\textbf{df}}} & \multicolumn{1}{l|}{\multirow{2}{*}{\textbf{F}}} & \multicolumn{1}{l|}{\multirow{2}{*}{\textbf{Sig.}}} & \multirow{2}{*}{\textbf{$\eta^2$}} \\ \cline{2-5} \cline{10-13}
 & \multicolumn{1}{l|}{M} & \multicolumn{1}{l|}{SE} & \multicolumn{1}{l|}{M} & \multicolumn{1}{l|}{SE} & \multicolumn{1}{l|}{} & \multicolumn{1}{l|}{} & \multicolumn{1}{l|}{} &  & \multicolumn{1}{l|}{M} & \multicolumn{1}{l|}{SE} & \multicolumn{1}{l|}{M} & \multicolumn{1}{l|}{SE} & \multicolumn{1}{l|}{} & \multicolumn{1}{l|}{} & \multicolumn{1}{l|}{} &  \\ \hline
Fidelity & \multicolumn{1}{l|}{3.23} & \multicolumn{1}{l|}{0.20} & \multicolumn{1}{l|}{3.71} & \multicolumn{1}{l|}{0.21} & \multicolumn{1}{l|}{(1,46)} & \multicolumn{1}{l|}{2.71} & \multicolumn{1}{l|}{0.107} & 0.057 & \multicolumn{1}{l|}{3.31} & \multicolumn{1}{l|}{0.19} & \multicolumn{1}{l|}{3.62} & \multicolumn{1}{l|}{0.20} & \multicolumn{1}{l|}{(3,44)} & \multicolumn{1}{l|}{3.67} & \multicolumn{1}{l|}{0.019*} & 0.204 \\ \hline
Interpretability & \multicolumn{1}{l|}{3.63} & \multicolumn{1}{l|}{0l19} & \multicolumn{1}{l|}{2.56} & \multicolumn{1}{l|}{0.19} & \multicolumn{1}{l|}{(1,46)} & \multicolumn{1}{l|}{15.30} & \multicolumn{1}{l|}{\textless{}.001*} & 0.254 & \multicolumn{1}{l|}{3.73} & \multicolumn{1}{l|}{0.17} & \multicolumn{1}{l|}{2.45} & \multicolumn{1}{l|}{0.18} & \multicolumn{1}{l|}{(3,44)} & \multicolumn{1}{l|}{9.84} & \multicolumn{1}{l|}{\textless{}.001*} & 0.407 \\ \hline
\end{tabular}%
}
\caption{ANOVA and ANCOVA (i.e., including curiosity and trust as covariates) results for the parking fines case. When controlling for curiosity and trust, both effects on fidelity and \chdeleted[id=del]{interoperability}\chadded[id=add]{interpretability} were significant. The effect on interpretability was also significant regardless of the covariates.}
\label{tab:main-effect-parking}
\end{table}

\subsubsection{Loan approval domain}

Results for the loan-approval domain are detailed in Table~\ref{tab:main-effect-laon}.
For the loan-approval case, an \chdeleted[id=del]{Analysis of Variance (}ANOVA\chdeleted[id=del]{)}  \chadded[id=add]{analysis} revealed that for the dependent variable of fidelity, there was no significant effect, F(1,46) = 1.97, p = .16, $\eta^2$ = .0542. Likewise, for the dependent variable of interpretability, the effect was statistically insignificant, F(1, 46) = 0.058, p = .81, $\eta^2$ = .001, suggesting that the particular type of knowledge manipulation that was conducted in this case has no significant effect on either fidelity or interpretability when not controlling for any covariate.

\chdeleted[id=del]{An Analysis of Covariance (ANCOVA) was conducted next to evaluate the effects of knowledge manipulation on the two dependent variables while controlling for the covariates trust and curiosity. The}\chadded[id=add]{ANCOVA} analysis revealed that for the dependent variable fidelity, the adjusted model was statistically significant, F(3, 44) = 19.16, p < .001,
$\eta^2$ = .572, indicating a substantial effect of the manipulation. Similarly, regarding the dependent variable interpretability, the adjusted model was also significant, F(3, 44) = 10.56, p = <.001, $\eta^2$ = .424, also showing a substantial effect. Thus, when incorporating trust and curiosity as covariates, the manipulation of knowledge type as was realized in the loan-approval domain, significantly affected the perceptions of fidelity and \chdeleted[id=del]{interoperability}\chadded[id=add]{interpretability} while controlling for trust and curiosity. 

Running the ANCOVA analysis also revealed that curiosity has a significant direct effect on the perception of fidelity, F(1,46) = 6.67, p = 0.013, $\eta^2$ = 0.13, and as in the two other domains, trust has a significant effect on both fidelity, F(1,46) = 51.40, p < .001, $\eta^2$ = .54, and on interpretability, F(1,46) = 31.58, p < .001, $\eta^2$ = .42.

Revisiting the related hypotheses (H1.3; H2.3), the results support the expectation that adding causal-related knowledge to process knowledge as an input to the LLM can potentially yield a significant effect on both perceptions of fidelity and \chdeleted[id=del]{interoperability}\chadded[id=add]{interpretability}. The effect on both is moderated by the user's curiosity about the explored condition and trust in the LLM tool. Beyond our original hypotheses, it is also observed that curiosity has a direct effect on fidelity, and that trust has a direct effect on both perceptions, interpretability and fidelity. 

\begin{table}[ht]
\centering
\def\arraystretch{1.2}%
\Large
\resizebox{\textwidth}{!}{%
\begin{tabular}{|l|llllllll|llllllll|}
\hline
\multirow{3}{*}{\textbf{Construct}} & \multicolumn{8}{l|}{\textbf{ANOVA}} & \multicolumn{8}{l|}{\textbf{ANCOVA}} \\ \cline{2-17} 
 & \multicolumn{2}{l|}{\textbf{Group 1}} & \multicolumn{2}{l|}{\textbf{Group 2}} & \multicolumn{1}{l|}{\multirow{2}{*}{\textbf{df}}} & \multicolumn{1}{l|}{\multirow{2}{*}{\textbf{F}}} & \multicolumn{1}{l|}{\multirow{2}{*}{\textbf{Sig.}}} & \multirow{2}{*}{\textbf{$\eta^2$}} & \multicolumn{2}{l|}{\textbf{Group 1}} & \multicolumn{2}{l|}{\textbf{Group 2}} & \multicolumn{1}{l|}{\multirow{2}{*}{\textbf{df}}} & \multicolumn{1}{l|}{\multirow{2}{*}{\textbf{F}}} & \multicolumn{1}{l|}{\multirow{2}{*}{\textbf{Sig.}}} & \multirow{2}{*}{\textbf{$\eta^2$}} \\ \cline{2-5} \cline{10-13}
 & \multicolumn{1}{l|}{M} & \multicolumn{1}{l|}{SE} & \multicolumn{1}{l|}{M} & \multicolumn{1}{l|}{SE} & \multicolumn{1}{l|}{} & \multicolumn{1}{l|}{} & \multicolumn{1}{l|}{} &  & \multicolumn{1}{l|}{M} & \multicolumn{1}{l|}{SE} & \multicolumn{1}{l|}{M} & \multicolumn{1}{l|}{SE} & \multicolumn{1}{l|}{} & \multicolumn{1}{l|}{} & \multicolumn{1}{l|}{} &  \\ \hline
Fidelity & \multicolumn{1}{l|}{4.05} & \multicolumn{1}{l|}{0.21} & \multicolumn{1}{l|}{3.62} & \multicolumn{1}{l|}{0.21} & \multicolumn{1}{l|}{(1,46)} & \multicolumn{1}{l|}{1.97} & \multicolumn{1}{l|}{0.16} & 0.042 & \multicolumn{1}{l|}{4.02} & \multicolumn{1}{l|}{0.14} & \multicolumn{1}{l|}{3.65} & \multicolumn{1}{l|}{0.14} & \multicolumn{1}{l|}{(3,44)} & \multicolumn{1}{l|}{19.16} & \multicolumn{1}{l|}{\textless{}.001*} & 0.572 \\ \hline
Interpretability & \multicolumn{1}{l|}{3.31} & \multicolumn{1}{l|}{0.22} & \multicolumn{1}{l|}{3.25} & \multicolumn{1}{l|}{0.23} & \multicolumn{1}{l|}{(1,46)} & \multicolumn{1}{l|}{0.058} & \multicolumn{1}{l|}{0.81} & 0.001 & \multicolumn{1}{l|}{3.27} & \multicolumn{1}{l|}{0.17} & \multicolumn{1}{l|}{3.31} & \multicolumn{1}{l|}{0.18} & \multicolumn{1}{l|}{(3,44)} & \multicolumn{1}{l|}{10.56} & \multicolumn{1}{l|}{\textless{}.001*} & 0.424 \\ \hline
\end{tabular}%
}
\caption{ANOVA and ANCOVA (i.e., including curiosity and trust as covariates) results for the loan approval case. When controlling for curiosity and trust,
both effects on fidelity and \chdeleted[id=del]{interoperability}\chadded[id=add]{interpretability} were significant.}
\label{tab:main-effect-laon}
\end{table}

\subsection{Direction of the effect}

\begin{figure}
    \centering
    \includegraphics[width=1\linewidth]{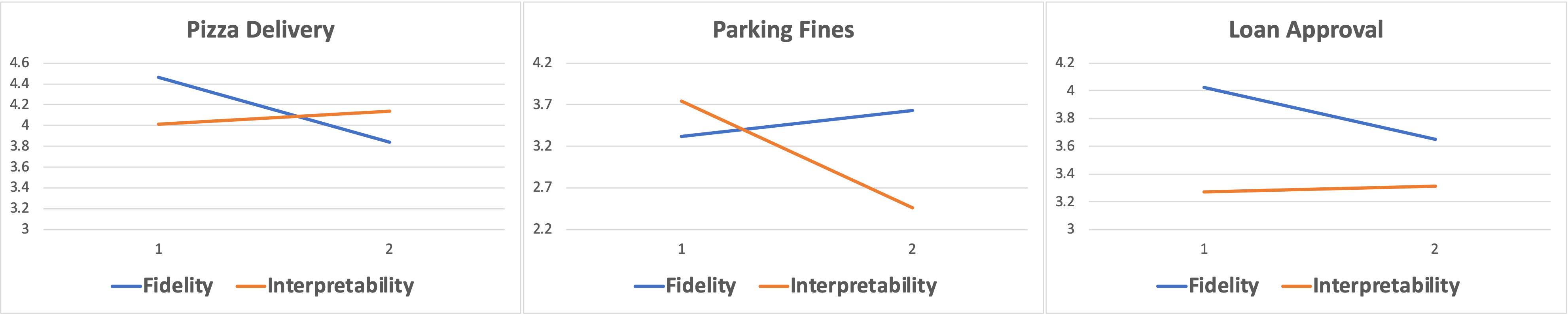}
    \caption{Constructs interaction}
    \label{fig:constructs-interaction}
\end{figure}

Careful observation of the study results revealed that, regardless of the significance of the effects, an important outcome that stands out across all three domains — one which we did not fully anticipate in our hypotheses — is related to the \textit{direction} of the effects on fidelity and \chdeleted[id=del]{interoperability}\chadded[id=add]{interpretability}. Our original thought was that informing the LLM prompt with additional knowledge, whether process-related (as in the pizza domain) or causal-related (as in the two other domains), would lead to a consistent effect direction on both dependent constructs, potentially improving both. However, this was, in fact, not the result. The actual effect directions, as illustrated in Figure~\ref{fig:constructs-interaction}, not only demonstrate an interaction between the two constructs (supported by prior literature~\cite{Markus2021}) but also show that the addition of either type of knowledge ingredient (process or causal) works towards improving the perceived fidelity of the explanation while compensating for the perceived interpretability of the explanation. Concerning the size of the effect, in the pizza delivery and loan approval domains, the improvement in fidelity was larger than the loss in interpretability. However, in the parking fines domain, the opposite was true. Hence, we highlight this particular caveat to be watched for in specific problem domains where one might attempt a similar type of intervention to `improve' the perceived quality of generated explanations.

\section*{\chdeleted[id=delr2]{9. Related Work}}



\chdeleted[id=delr2]{This study builds on our previously published extended abstract~\cite{Fahland2023WhyLate} where the original idea of synthesizing different knowledge ingredients for improved BP explanations given by LLMs was conceived. 
In this work, we further expand from these roots in various directions. First, the core idea is formalized in a set of hypotheses. Second, a library of services has been developed to automate the concept and integration with LLMs. Third, to evaluate the hypotheses, we rigorously developed a corresponding scale to assess the effectiveness of the altered inputs and employed it to test the effectiveness of using the services with GPT4.0 and the reliability of the scale itself. This effort is followed by a quantitative analysis of user responses and derived conclusions.}



\chdeleted[id=delr2]{In the following, we discuss how our study relates to and contributes to the literature on 
the intersection between LLMs and BPs for the sake of BP explainability.} 


\chdeleted[id=delr2]{Overall, the use of LLMs for BPs has been researched only very recently, enabled by the availability and accessibility of the GPT-based foundation models. In a broader perspective, BP characteristics present a unique opportunity for the development of a new class of foundation models that utilize the timely sequencing of processes for various process-related tasks such as activity prediction, process optimization, and decision making (e.g.,~\cite{Beheshti2023ProcessGPT:Intelligence}~\cite{Rizk2024}).
Vigdof et al.~\cite{vidgof2023large} explore the potential and broader application for integrating LLMs along all stages of the BPM life-cycle and propose further broad research directions for the use of LLMs in BPM. While covering process mining, optimization, and decision-making, the paper does not conduct experiments. Grohs et al.~\cite{Grohs2024} demonstrate in several experiments that advanced LLMs such as GPT4.0 are capable of transforming natural language descriptions of smaller processes into declarative and imperative process models and can aid in detecting automate-able tasks from natural language process descriptions. None of the above papers delves into the detailed use of LLMs for explainability in BPs nor examines the consequences of such employment, which is the focus of our study. Berti et al.~\cite{Berti2024} demonstrate the ability to infer process abstractions from event data and subsequently feed these to an LLM along with a query answering process-related questions. While there may be some overlap in eliciting the process view as input for the LLM, our work extends the input to the LLM by incorporating causal and XAI views to derive sound and interpretable explanations.}


\chdeleted[id=delr2]{Prompt engineering for LLMs that are employed towards various tasks in BPM is also recently surveyed in~\cite{Rizk2024}. While zero-short or few-short learning is reported to improve the performance of the LLMs, other techniques such as modifying its structure, and even subtle reordering of the various input elements, may have destructive effects on the performance of the LLM. These experiences have recently also resulted in a public repository of prompts~\cite{Bach2022}. To mitigate such destructive consequences, Jessen et al.~\cite{Jessen2023} propose a fully automated, structured process for generating prompts for translating natural-language analysis questions into SQL queries over event data. Their approach relies on (iteratively) enriching the prompt with additional information from different perspectives, which then results in a more stable quality of answers (correct SQL queries). While our approach for prompt engineering bears similarities as we also integrate information from multiple perspectives, our focus is on altering the content of the input (semantics) to the LLM rather than reshuffling its structure (syntax). We pursue this to address BP explainability and leverage information in a knowledge graph to automatically generate the prompt.}

\chdeleted[id=delr2]{A general assumption made by existing studies on LLMs for BPM is that the textual output generated by an LLM is inherently understandable and suitable for the user. While some prior studies do analyze fidelity~\cite{Jessen2023}, this study is the first to also investigate the interpretability of LLM output by users and other background factors that influence these. Notably, we establish that users perceive a trade-off between fidelity and interpretability.}

\chdeleted[id=delr2]{Several works studied the application of explainability techniques for process prediction and prescription. For example, Shapley Values~\cite{shapley:book1952} have been used to explain which features influence the prediction made by an ML model learned from an event log~\cite{GalantiCLCN2}. Shapley values are also used as a feature for generating explanations for the recommendations produced by a prescriptive model~\cite{Padella2022}. Several other works also explore (combinations of) other ``explainability'' techniques for process prediction and recommendation, e.g.,~\cite{El-Khawaga2022},~\cite{Stevens2021},~\cite{Velmurugan2021},~\cite{Wickramanayake2023}. Generally, this line of research focuses on explaining a predictive or prescriptive model of a process, i.e., it aims to explain how an approximation of the observed event log relates input to output features. We extend the conventional applicability of XAI techniques to BP explainability to also include other perspectives such as causal and process views which are inferred directly from the process event log.}

\chdeleted[id=delr2]{Further, all studies have in common that they generate explanations in the form of charts that visualize the influence of input on output features. A user study confirmed that the generated visualizations ``are generally comprehensible to correctly carry out analysis’ tasks'' for prediction~\cite{Galanti2023AnAnalytics}. The study also observed that accuracy in understanding influences on outcomes and the overall quality of the approach (esp. on difficult tasks) could improve when analysts are better aided in understanding how different influencing factors integrate towards the final effect.  Another user study~\cite{Rizzi2022} showed that also experts undergo a learning curve when interpreting the plots and require additional explanations for answering more difficult tasks. Our study contributes to integrating various features for explanations and complements these prior studies by analyzing the factors that influence the understandability of explanations in natural language.}

\chdeleted[id=delr2]{A novel component in this work is the incorporation of the causal execution view as a basis for generating high-fidelity explanations. With regards to the applicability of causal inference and discovery to BPs, recently, Dasht Bozorgi et al. demonstrated how to apply statistical causality analysis techniques to identify cause-effect relations in BPs from event logs~\cite{Bozorgi2020}. These causality analysis techniques can be used to improve control policies using reinforcement learning~\cite{Bozorgi2021PrescriptiveReduction} or to improve prescriptive process monitoring by estimating the causal effect of an intervention on a performance metric~\cite{DashtBozorgi2023PrescriptiveEstimation}. 
While Dasht Bozorgi et al. focus on identifying individual rules of cause-effect of an action utilizing the Conditional Average Treatment Effect (CATE) technique~\cite{Kunzel2019MetalearnersLearning}, our work employs a novel approach that utilizes the timing of the activities for the discovery of causal execution interdependency between process activities, and use this view as an additional input to the LLM.}

\section{\chadded[id=addr3]{Study Limitations}}
\label{sec:limitations}


\chadded[id=addr2]{
As with the operationalization of any empirical study, our work is not without limitations.}

\chadded[id=addr2]{First, to ensure the internal validity of our developed scale concerning the content manipulation of explanations presented in our experiment, we deemed it necessary to either control or measure for additional undesired factors that could interfere with the results. To address this, we chose to measure perceptions of trust and curiosity as two potential covariates, while utilizing the best-in-class LLM currently known. This approach ensured that our developed scale remained sensitive to capturing the perceived quality of the explanations' content.}
\chdeleted[id=addr2]{We understand that once we have introduced the instrumentation, it calls for employing our developed scale for LLM benchmarking. A comparison between different LLMs is a promising future direction to be pursued, now made feasible through the availability of our developed instrumentation to facilitate such benchmarking.}

\chadded[id=addr2]{Second, recognizing users' perception of explanation quality as a key factor influencing trust and usage intention in systems generating explanations, we focused \chadded[id=addr3]{solely} on developing an instrument aimed at assessing the perceived quality rather than the "objective" quality of the explanation. This focus necessitated accounting for the latter. To control for the potential influence of the true quality of the explanation, our approach involved both the study researchers (i.e., the authors) reviewing the generated narratives to ensure their fidelity to the ground truth and presenting the ground truth explanations to participants. Regarding the former, while the researchers are experts in BPM, we acknowledge that \chadded[id=addr3]{objective fidelity metrics (e.g., precision and recall) could have been employed using a textual articulation of all knowledge ingredients about the underlying business process as a basis for assessment. In addition,} a more rigorous methodology could have been employed, such as utilizing a panel of experts or leveraging an LLM-as-a-judge, or extending the survey to include additional explicit measurements of participants’ perceptions of the explanations' correctness relative to the ground truth. The results could then have been normalized based on these ratings. As for the latter, presenting the ground truth explanations ensured that participants’ ratings were always relative to their perception of the explanations’ true correctness.}


\chadded[id=addr3]{Regarding potential threats to the reproducibility and generalizability of our results, we addressed the former by developing a rigorous instrumentation process and a standardized user study procedure, making our materials publicly available. However, we acknowledge that the latter may raise concerns due to the limited diversity of LLMs employed. While, in principle, this aspect is governed by the nature of explanation embodiment—ultimately articulated in natural language (English)—this threat could be further mitigated by increasing the diversity of LLM models, including open source alternatives, and conducting comparative benchmarking among them.} \chadded[id=addr2]{A comparison between different LLMs is a promising future direction to be pursued, now made feasible through the availability of our developed instrumentation to facilitate such benchmarking.}

\section{Conclusions and Future Work}
\label{sec:conclusions}

\chadded[id=addr2]{This study builds on our previously published extended abstract~\cite{Fahland2023WhyLate} where the original idea of synthesizing different knowledge ingredients for improved BP explanations given by LLMs was conceived. 
In this work, we further expand from these roots in various directions. First, the core idea is formalized in a set of hypotheses. Second, a library of services has been developed to automate the concept and integration with LLMs. Third, to evaluate the hypotheses, we rigorously developed a corresponding scale to assess the effectiveness of the altered inputs and employed it to test the effectiveness of using the services with \chadded[id=add]{GPT4.0}\chdeleted[id=del]{ChatGPT 4}, and the reliability of the scale itself. This effort is followed by a quantitative analysis of user responses and derived conclusions.}

\chdeleted[id=del]{In regards to}\chadded[id=add]{Regarding} the hypotheses, our results clearly indicate that providing the LLM with additional knowledge about causal execution dependencies, or even just the temporal sequencing embedded in the process knowledge, can enhance the perceived fidelity. 
However, this effect may diminish entirely when the user has low trust in the LLM. It may also lessen when the user has limited curiosity about the problem for which the explanation is sought. One should also exercise caution in overly emphasizing the potential benefit of increased fidelity, as it may come at the expense of reduced perceived interpretability. Further examination of the trade-off between fidelity and \chdeleted[id=del]{interoperability}\chadded[id=add]{interpretability}, and the extent to which an explanation remains ``sufficient'' may be explored in the future.

For pragmatic reasons, we limited the number of manipulations. Therefore, we focused on those closely aligned with the novelty of incorporating the causal knowledge ingredient with the process one for explainability. We also explored how the more conventional approach of using XAI in itself is influenced when integrated with the process perspective. Additionally, we investigated whether incorporating the causal view enhances the perceived quality when combined with the process component. We acknowledge that exploring the effects of augmenting process knowledge with XAI, as opposed to using only process knowledge, may still be an interesting direction to pursue. In general, other process-related perspectives could also be incorporated as additional knowledge ingredients.

We merged the manipulation of knowledge components with that of the domain. We have made all experimental instruments available to facilitate future work in replicating our methods and extending the examination to other problem domains and communities.

The generation of knowledge-informed explanations could be fully automated with the use of the tooling framework developed in this paper. Additional `tweaking' of the prompt could be employed to mitigate the potential drawback with interpretability, e.g., by embedding prompt instructions to the LLM to also try to summarise the result as much as possible without compromising fidelity. Statements such as ``make the explanation not longer than 2-3 sentences without losing on its core detail, correctness, and accuracy.'' may be attempted. We already have the result of such a feature illustrated in figure~\ref{fig:parking-fines-screenshot}. However, we did not include it in the experiment to keep the output of the LLM conform to its default hyper-parameter configuration, keeping the configuration consistent across all manipulations (e.g., temperature, top p, max length).

\chdeleted[id=addr3]{
As with the operationalization of any empirical work, our study is not without its limitations.}

\chdeleted[id=addr3]{First, to ensure the internal validity of our developed scale concerning the content manipulation of explanations presented in our experiment, we deemed it necessary to either control or measure for additional undesired factors that could interfere with the results. To address this, we chose to measure perceptions of trust and curiosity as two potential covariates, while utilizing the best-in-class LLM currently known. This approach ensured that our developed scale remained sensitive to capturing the perceived quality of the explanations' content.
We understand that once we have introduced the instrumentation, it calls for employing our developed scale for LLM benchmarking. A comparison between different LLMs is a promising future direction to be pursued, now made feasible through the availability of our developed instrumentation to facilitate such benchmarking.}

\chdeleted[id=addr3]{Second, recognizing users' perception of explanation quality as a key factor influencing trust and usage intention in systems generating explanations, we focused on developing an instrument aimed at assessing the perceived quality rather than the "objective" quality of the explanation. This focus necessitated accounting for the latter. To control for the potential influence of the true quality of the explanation, our approach involved both the study researchers (i.e., the authors) reviewing the generated narratives to ensure their fidelity to the ground truth and presenting the ground truth explanations to participants. Regarding the former, while the researchers are experts in BPM, we acknowledge that a more rigorous methodology could have been employed, such as utilizing a panel of experts or leveraging an LLM-as-a-judge, or extending the survey to include additional explicit measurements of participants’ perceptions of the explanations' correctness relative to the ground truth. The results could then have been normalized based on these ratings. As for the latter, presenting the ground truth explanations ensured that participants’ ratings were always relative to their perception of the explanations’ true correctness.}

\chadded[id=addr3]{Aside from extending the diversity of LLMs employed as discussed above,} \chadded[id=addr2]{future work may include additional directions. }\chdeleted[id=delr2]{We note that}
\chadded[id=add]{Our assessment of explanations' quality, as generated by LLMs in this work, conforms to the philosophy of LLMs as a means to synthesize and articulate the output of the system, while the content underlying such explanations is computed from the process event logs by employing state-of-the-art techniques that are assumed to retain its correctness. Future work may delve further into evaluating the potential use of LLMs as a means to infer the content for such explanations, where assessing the quality of such explanations may require a quantitative measurement scale to assess their correctness with respect to the underlying process recorded in the log.}

\chadded[id=addr2]{Another direction}\chdeleted[id=delr2]{Future work} might include the templating of the prompts as a means to fully automate the blending of contemporary techniques for prompt engineering (e.g., sequencing and restructuring) and our content manipulation to combine all as an input to the LLM. 
Our use of the LLM in this work relied on a one-off, out-of-the-box execution of \chadded[id=add]{GPT4.0}\chdeleted[id=del]{ChatGPT-4} with no further model tuning (i.e., few-shot learning) to train the language model to interpret the input narratives, with a focus on explanations that arise from causal, process, and XAI knowledge synthesize.

Next steps could leverage \chdeleted[id=del]{on} our developed scale to \chdeleted[id=del]{methodological}\chadded[id=add]{methodically} gather \chdeleted[id=del]{scoring}\chadded[id=add]{scores} for process explanations and subsequently use these scores to prioritize among a ``repository'' of explanations to fine-tune the LLM model.



\bibliographystyle{elsarticle-num} 
\bibliography{SAX-references}


\end{document}